# FormuLLA : A Large Language Model Approach to Generating Novel 3D Printable Formulations

Adeshola Okubena[1], Yusuf Ali Mohammed[1] and Moe Elbadawi[1].*

[1]School of Biological and Behavioural Sciences, Queen Mary University of London, Mile End Road, London E1 4DQ, UK.

* Corresponding Author: m.elbadawi@qmul.ac.uk

**Abstract**

Pharmaceutical three-dimensional (3D) printing is an advanced fabrication technology with the potential to enable truly personalised dosage forms. Recent studies have integrated artificial intelligence (AI) to accelerate formulation and process development, drastically transforming current approaches to pharmaceutical 3D printing. To date, most AI-driven efforts remain narrowly focused, while failing to account for the broader formulation challenges inherent to the technology. Recent advances in AI have introduced artificial general intelligence concepts, wherein systems extend beyond conventional predictive modelling toward more generalised, human-like reasoning. In this work, we investigate the application of large language models (LLMs), fine-tuned on a fused deposition modelling (FDM) dataset comprising over 1400 formulations, to recommend suitable excipients based on active pharmaceutical ingredient (API) dose, and predict filament mechanical properties. Four LLM architectures were fine-tuned, with systematic evaluation of both fine-tuning and generative parameter configurations. Our results demonstrate that Llama2 was best suited for recommending excipients for FDM formulations. Additionally, model selection and parameterisation significantly influence performance, with smaller LLMs exhibiting instances of 'catastrophic forgetting'. Furthermore, we demonstrate: (i) even with relatively small dataset of over 1400 formulations, it can lead to model 'catastrophic forgetting'; (ii) standard LLM metrics only evaluate linguistic performance but not formulation processability; and (iii) LLMs trained on biomedically-related data do not always produce the best results. Addressing these challenges is essential to advancing LLMs beyond linguistic proficiency and toward reliable systems for pharmaceutical formulation development.

### Graphical Abstract

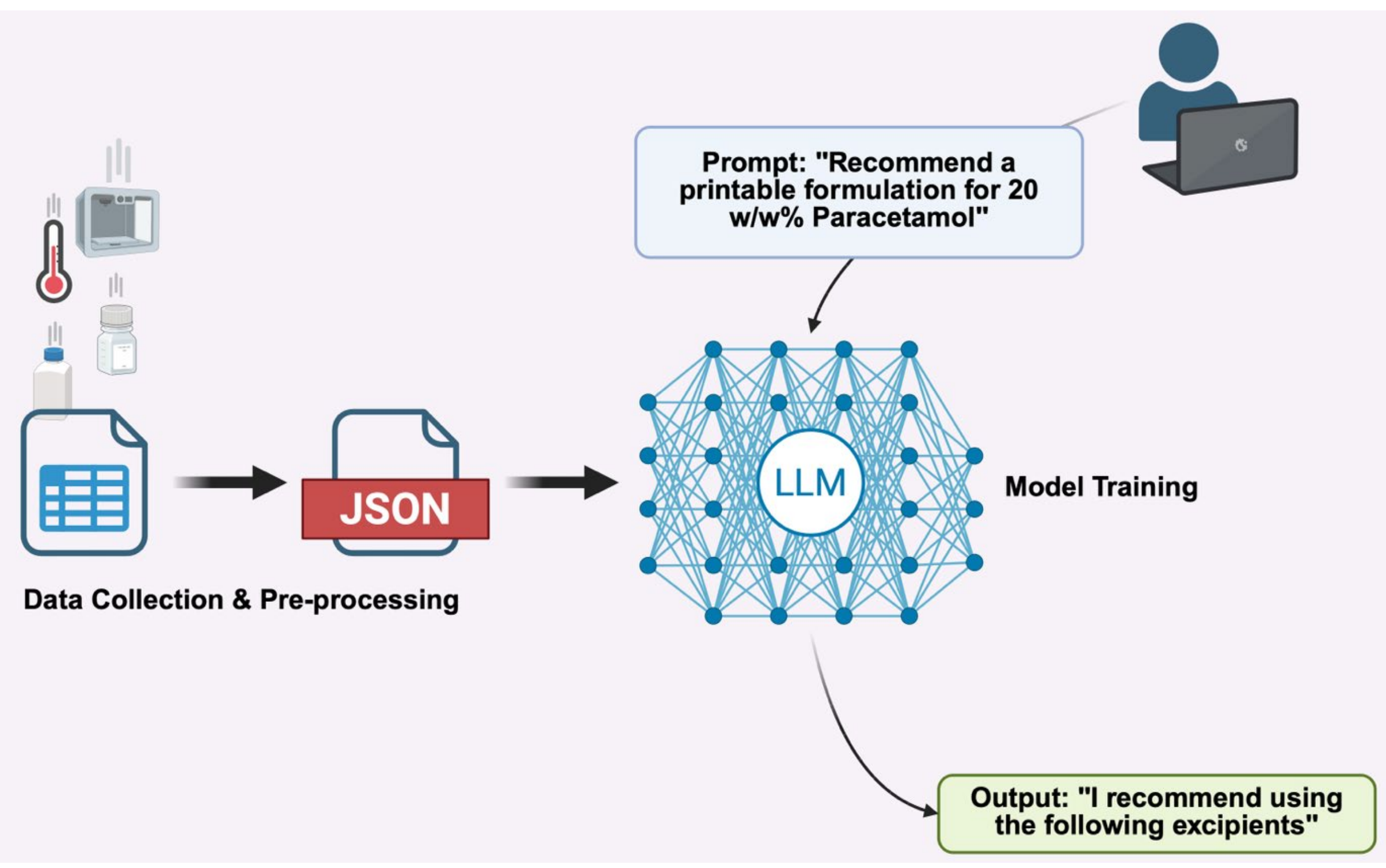

### List of Abbreviations

| Keyword | Definitions |
|---|---|
| APIs | Active Pharmaceutical Ingredients |
| AI | Artificial Intelligence |
| ML | Machine Learning |
| DL | Deep Learning |
| LLMs | Large Language Models |
| PEFT | Parameter-efficient Fine-tuning |
| LoRA | Low-rank adaptation |
| BLEU | Bilingual Evaluation Understudy |
| ROUGE | Recall-Oriented Understudy for Gisting Evaluation |
| VELVET | Validation Excipient List Verification and Evaluation Tool |
| ANOVA | Analysis of Variance |
| SFT | Supervised Fine Tuning |

# 1 Introduction

Three-dimensional (3D) printing, or additive manufacturing, has emerged as a transformative technology with the potential to revolutionise multiple manufacturing sectors [1-5]. In the pharmaceutical domain, 3D printing enables the precise fabrication of complex drug delivery systems and customised dosage forms tailored to individual patient needs [6-8]. This capability positions 3D printing as a strong contender for advancing personalised medicine, providing the potential for enhanced therapeutic efficacy, reducing adverse effects, and improving patient adherence. However, despite its promise, the development of 3D-printed medicines remains at a relatively early stage. The process of designing viable formulations still depends heavily on empirical experimentation, making it resource-intensive and time-consuming [9, 10].

To accelerate progress in this area, computational approaches have been increasingly adopted to support and complement laboratory-based formulation design. Artificial intelligence (AI) has proven particularly effective in this regard, offering the ability to model complex systems, identify hidden correlations, and predict formulation performance based on large datasets. Traditional machine learning (ML) techniques have achieved notable success in tasks such as predicting solubility, dissolution rates, and processability [11, 12]. Yet, these methods are primarily discriminative as they classify or predict based on existing data rather than generating new, innovative formulations. To address this limitation, generative models such as conditional Generative Adversarial Networks (cGANs) have been investigated. Previous studies have demonstrated the potential of cGANs to generate de novo formulations by learning from existing datasets [13]. cGANs are neural networks that consist of two competing networks, a generator that produces synthetic data and a discriminator that evaluates its authenticity. Through iterative adversarial training, the Generator learns to create increasingly realistic samples. While effective in certain contexts, cGANs often require carefully engineered input representations and are prone to unstable convergence. Moreover, they struggle to integrate unstructured or text-based pharmaceutical knowledge, limiting their utility in data-sparse or heterogeneous formulation environments. Ultimately, whether it is traditional discriminative ML or generative models like cGANs, they are constrained to a narrow task, sometimes even a single task.

Large Language Models (LLMs) are an emerging group of generative models that are revolutionising multiple sectors [14-19]. Their core neural network architecture, the transformer [20], addresses traditional generative bottlenecks such as long-range dependency handling and limited parallelisation, that has allowed them to be applied across numerous tasks. This design allows LLMs to capture contextual relationships over extended sequences while enabling efficient, large-scale training through parallel processing. As a result, LLMs not only achieve superior contextual understanding compared to earlier recurrent or convolutional architectures but also exhibit exceptional scalability and adaptability across domains. These characteristics make them uniquely capable of learning complex, high-dimensional relationships between inputs, such as the interplay between chemical structures, formulation components, and functional properties in pharmaceutical systems. offer a fundamentally different generative framework, one grounded in sequence modelling rather than adversarial optimisation. Trained on vast textual corpora, LLMs learn contextual and semantic relationships between words, enabling them to generate coherent, knowledge-informed outputs across a variety of domains. Unlike cGANs, which rely on adversarial feedback to mimic data distributions, LLMs use next-token prediction to capture both syntactic and conceptual relationships, making them well-suited for domains like pharmaceutics where information exists in mixed formats (e.g., numerical data, chemical descriptions, and textual records).

In addition to their enhanced pattern recognition, LLMs are widely adopted because users can interact with an ML model using human language and can have an interactive dialogue with the model. This is in contrast to previous ML work where the model were narrowly focused on a specific task and one can have a severely limited dialogue. For example, in training an LLM, users can have it learn specific patterns and then have a discussion with the model in and around the topic [21]. This is analogous to a scientific expert learning a new instrument (e.g., electron diffraction) and then users can ask the expert questions about the instrument (e.g. how does it work?) but around the subject (how is it suitable for analysing excipients?). In contrast, traditional ML techniques are more analogous to an “expert” trained solely to perform a single task, such as predicting whether a sample is appropriately prepared for the instrument.

The success in LLM can be observed across a number of fields. For example, in drug discovery, LLMs have been found to comprehend multi-modal data for accelerating discovery and repurposing [22-26]. In material science, they have been found to discover novel materials, facilitate material fabrication and expertly analyse data [27]. Despite these successes, their use in drug formulation development is yet to be explored.

LLMs are demonstrably larger than conventional ML models used in pharmaceutics. Aware of this, methods have been developed to leverage existing LLMs, already pre-trained on copious amount of data, and repurpose their use in a domain-specific application. One such method is referred to as parameter efficient fine tuning (PEFT), where only a fraction of the, potentially, billions of parameters are adjusted on a domain specific dataset [28]. PEFT extends this potential by adapting a pre-trained LLM to a specific task or domain using a smaller, targeted dataset. Through methods such as supervised fine-tuning (SFT) or parameter-efficient approaches (e.g., LoRA adapters), an LLM can be refined to perform domain-specific reasoning, such as predicting suitable excipient combinations for a given active pharmaceutical ingredient (API). This represents a conceptual shift from cGAN-based generative modelling to knowledge-grounded reasoning, where the model leverages both linguistic understanding and contextual pattern recognition to make informed formulation suggestions.

In this study, we investigated the effect of PEFT LLMs on pharmaceutical 3D printing formulation data. We experimented with four different LLMs, each with its unique architecture and/or training dataset. The performance of each model was assessed with established natural language processing metrics (e.g., BLEU, ROUGE-1, ROUGE-2 and ROUGE-L) and a custom metric designed to evaluate the accuracy of excipient recommendations for a given API. The overarching aim is to determine whether fine-tuned LLMs can generate reasonable excipient selections based on our fine-tuning training dataset and whether LLMs can be harnessed in formulation development.

# 2 Methods

## 2.1 Dataset and Preprocessing

The dataset acquired from Elbadawi et al. 2020 was formatted for machine learning [11, 12]. The dataset is a .csv that contains different excipients in each formulation row; when present, the drug was listed with the quantity in w/w%. This allowed the corresponding excipient mixture of each drug to be extracted, according to its quantity in the row of the API. The other useful values in the final dataset were the filament aspect value in that row, which represented whether that specific combination of API-excipients produced a brittle, good, “unextrudable”, or unknown formulation for 3D printing. Finally, the formulation's printability stated whether it was suitable for 3D printing.

The dataset was originally formatted for discriminative ML models, similar to a DoE format. The inputs for the discriminative ML were the formulation composition in w/w% and the output were the printability as a binary label ‘yes’ or ‘no’ and the mechanical aspect of the filament. In this process, the ML models are given hundreds of inputs about the composition and tasked with predicting two outcomes. The reverse process is well-known as challenging in the ML domain, whereby the inputs are a few but the outputs are many. However, this is more pragmatic in the pharmaceutical field, where researchers would prefer to ask the ML model given an API “give me the correct excipients”. This is the benefit LLMs but the dataset needs to be formatted correctly.

The dataset adhered to the Alpaca format for LLM training, which provides the LLMs a set of instructions and responses pairs, which are akin to ‘inputs’ and ‘outputs’ in discriminative ML models. For each formulation, a python pipeline was programmed to extract the API and corresponding proportion and placed into a column called ‘Instructions’. The same python programming extracted, per formulation, the excipient and their proportion, the printability and filament aspect – and inserted it into a column called ‘Response’. Then, for the dataset's labelling, a function was created to give the base model prompting in the Alpaca format. Therefore, prompts follow this structure: instruction, input, response, ending in an EOS token. As LLMs can comprehend language, a standardised text was used to contextualised both input and output, which is presented in Table 1. This new formatted dataset was saved as a JSONL format, which is common for LLM training. Then the function was mapped to the dataset and tokenised. This prepared the dataset for supervised fine-tuning.

*Table 1. Examples of instructions and responses used to fine-tune the LLMs.*

| **Instructions** | **Response** |
| --- | --- |
| “Recommend excipients for 20 w/w% Ciprofloxacin” | “For this formulation, use these excipients: PCL: 60 w/w%, PEG2000: 20 w/w%. This is printable and has a flexible filament aspect.” |
| “Recommend excipients for 10 w/w% Theophylline” | “For this formulation, use these excipients: HPC: 40 w/w%, Eudragit: 40 w/w%, PEG6000: 10 w/w%. This is printable and has a Good filament aspect.” |
| “Recommend excipients for 25 w/w% Paracetamol” | “For this formulation, use these excipients: HPMC: 60 w/w%, Methyl paraben: 10 w/w%, Polyethylene glycol PEG8000: 5 w/w%. This is printable and has a Good filament aspect.” |

## 2.2 Model Selection

There are many LLMs, which continue to increase. For this study, we focused on open-source LLMs, of which we identified four candidates, by which we mean models whose architectures and trained weights are publicly available and can be freely accessed, modified, and deployed by the research community under permissive licences. From this space, we identified four representative candidate models for further evaluation. The most popular LLM is Chat GPT, however, as it is not open-sourced, it was not investigated in this study.

**Llama 7B**

Large Language Model Meta AI (Llama) is developed by Meta AI and released as part of the LLaMA family of open-foundation models. LLaMA models are trained on a large mixture of publicly available and licensed datasets encompassing billions to trillions of tokens of text to capture broad language patterns. Essentially, the are trained on web and curated text data to support general language tasks. LLaMA 2 largely uses the same training dataset as the original Llama [29] and comes in multiple sizes (e.g., 7B, 13B, 70B), and due to computational limits we investigated 7B, which is shorthand for 7 billion parameters. The aim of Llama models is to focus on high performance whilst maintaining efficiency, with a view to making them open-sourced so they can be leveraged by the wider community [30].

**Mistral**

Mistral was developed focusing on efficient, open-weight LLMs. It is trained on diverse multi-lingual corpora. It is also designed to balance high-performance with efficiency, using grouped-query attention and sliding window attention. The LLM can tackle tasks in common reasoning, world knowledge, math, coding and reading comprehension [31].

**T5-XL**

T5 are developed by Google Research and are a text-to-text transfer transformer (T5). They are trained on curated data referred to as "Colossal Cleaned Common Crawl" (C4) [32], also curated by Google researchers. It is unique amongst the other LLMs because it follows an encoder-decoder architecture, whereas the rest are decoder-only transformers [33].

**BioGPT**

Unlike the above, BioGPT is a domain-specific LLM that follows the GPT-2 architecture. It is trained on 115 million PubMed abstracts, with the aim of enabling improved language tasks in the biomedical domain. It's been tasked with end-to-end relation extraction, question answering, document classification and text generation tasks, and generally performed better than GPT-2 in such tasks for biomedical applications. When asked about specific drugs, BioGPT was said to generate "more specific and professional descriptions." [34]

Table 2 summarises the LLMs, with respect to their model size, architecture and training dataset. As presented and discussed in this subsection, each LLM is unique.

*Table 2.The four LLMs experimented with and their respective properties. Note 'M' and 'B' indicates million and billion, respectively.*

| LLM | Model Size | Architecture | Training dataset |
|---|---|---|---|
| Llama2 | 7B | Decoder Transformer | Common Crawl, C4, GitHub, Wikipedia, Books, ArXiv, StackExchange |

| Mistral | ~7B | Decoder Transformer | Multi-lingual internet and curated datasets |
|---|---|---|---|
| T5-XL | ~3B | Encoder-Decoder Transformer | C4 |
| BioGPT | ~350M | Decoder Transformer | PubMed abstracts |

## 2.3 Training

Following data preparation, the dataset were split using the hold-out method into training and testing. For efficient, faster training and the use of less memory and storage, quantisation was used either by the quantised version of the LLM with the unsloth library (Llama 2 and Mistral) or quantisation of the full model with the bytes and bits library (BioGPT). However, for T5-XL, the full model was used as neither approach was compatible.

Instead of full fine-tuning, which updates all parameters of the base model, parameter-efficient fine-tuning (PEFT) was used to update only a small subset of trainable parameters [35]. Specifically, we employed Low-Rank Adaptation (LoRA), which injects trainable low-rank adapter matrices into selected layers of a frozen base model, enabling faster and more efficient training [36] (Figure 1). For each LLM, the baseline configuration applied LoRA to the attention Query (Q) and Value (V) projections. Additional experiments also included the Key (K) and Output (O) projections to assess whether expanding the set of adapted layers affected generation performance.

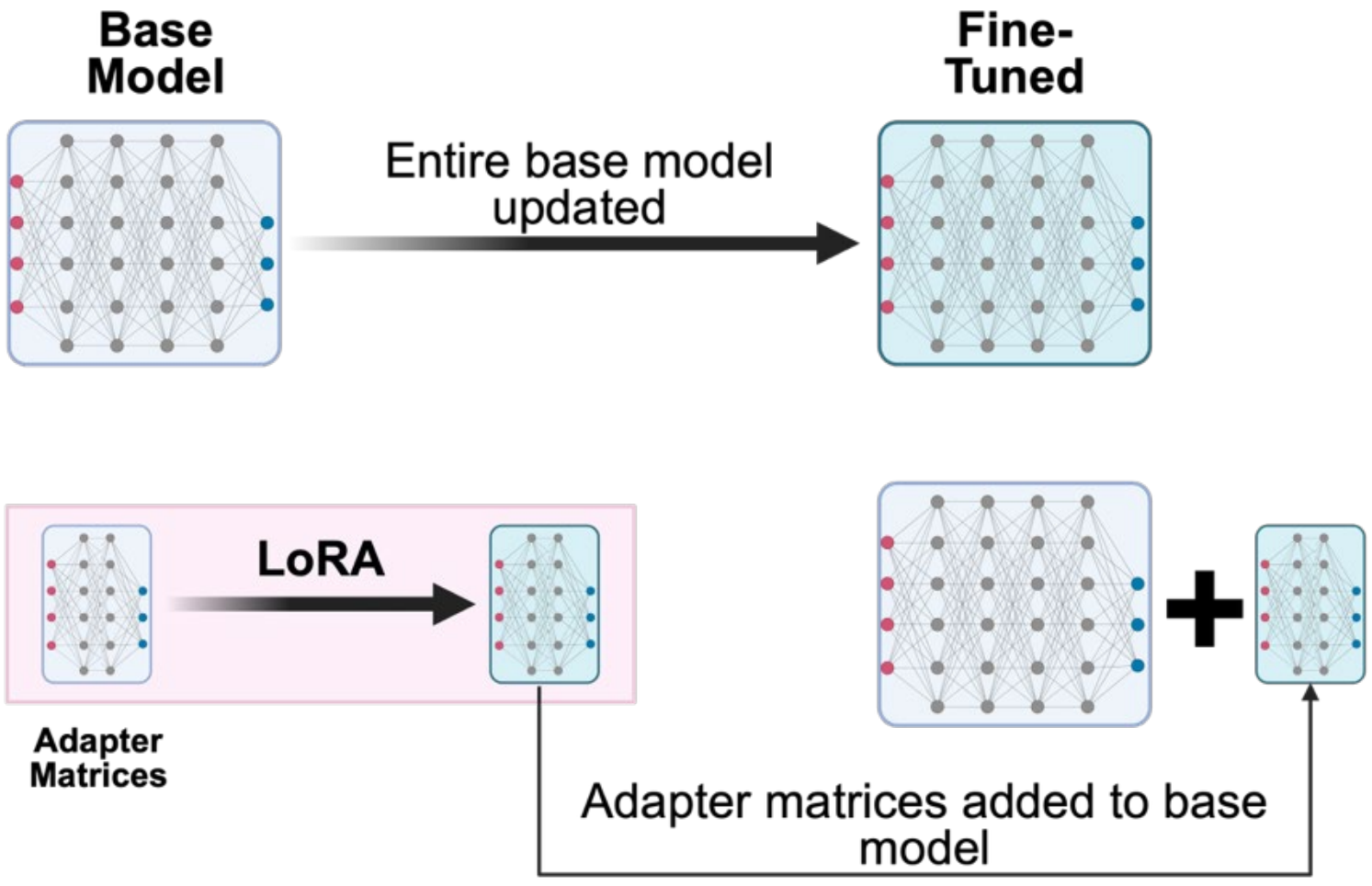

*Figure 1. Depiction of the difference between full fine-tuning and fine-tuning with LoRA. LoRA is more computationally efficient as it updates a small number of parameters, rather than the entire model.*

During the training process, an epoch loss logger was added to obtain the average loss per epoch. This was to serve as a preliminary model performance evaluator, which measures the difference between predictions and actual results (loss), indicating prediction accuracy. The larger the loss, the greater the difference between actual and predicted results; therefore, the LLM would have lesser predictive accuracy [37].

## 2.4 Hyperparameter tuning

To evaluate the impact of individual hyperparameters on model performance, baseline hyperparameters were chosen. These included a top-p value of 0.9, a temperature of 0.7, and a learning rate of $10^{-4}$, and targeted the 'Q' and 'V' LoRA layers for PEFT. These hyperparameters were designated as each LLM's "standard" condition. In each experiment, only one hyperparameter was varied at a time while the others remained fixed at their baseline values. This strategic approach allowed for isolated assessment of each hyperparameter's influence on evaluative metrics.

We further explored the effect of learning rate, with values of $10^{-2}$, $10^{-4}$, and $10^{-6}$ were tested across all four LLMs, resulting in 12 experiments. Moreover, the LoRA configuration was modified to include the addition of 'K' and 'O' layers in four experiments, making the total number of training hyperparameter experiments 16.

Separately, generation hyperparameters top-p and temperature were varied. The temperature parameter adjusts how confidently the model selects the next token, with lower values make the output more deterministic and focused, while higher values encourage more varied and creative generations. The top p (nucleus sampling) parameter, on the other hand, limits token selection to a dynamic subset of the vocabulary whose cumulative probability mass is below a threshold p, effectively filtering out unlikely candidates while retaining diversity. In essence they determine how "safe" or "adventurous" the model's responses are, balancing precision with creativity. As LLMs are generated for general applications, such parameters are beneficial in some sectors, such as poetry, storytelling, or dialogue generation, where creativity and variation are desired. However, in highly domain-specific contexts, such as pharmaceutical formulation or materials science, excessive randomness can introduce inconsistencies or chemically implausible combinations. Therefore, careful calibration of temperature and top p values is needed to maintain both diversity and domain fidelity. In other words, while these parameters can make a model sound more imaginative, they must be tightly controlled in technical fields to ensure the outputs remain scientifically valid and practically relevant.

Each LLM underwent 11 such experiments, totalling 44 across all models. Altogether, 60 experiments were conducted to systematically investigate the effects of training and generation hyperparameters across the selected LLMs (Table 3). Following this, the LLM was able to generate according to the test dataset, and line-by-line predictions were saved in a JSONL format alongside the input and the reference for comparison and further evaluation.

*Table 3. List of generation and training hyperparameters used and the individual values investigated, for each LLM. *When temperature or top-p is 0, the do sampling hyperparameter was set to false.*

| **Generation Hyperparameters** | | **Training Hyperparameters** | |
|---|---|---|---|
| **Top-p** | **Temperature** | **Learning Rate** | **Additional LoRA layers** |
| 0.9 | – | $1\times10^{-2}$ | Key + Output |
| 0.7 | 0.7 | $1\times10^{-4}$ | – |
| 0.5 | 0.5 | $1\times10^{-6}$ | – |
| 0.3 | 0.3 | – | – |
| 0.1 | 0.1 | – | – |
| 0.0* | 0.0* | – | – |

## 2.5 Evaluation

Standard natural language metrics were used, which were bilingual evaluation understudy (BLEU), and several recall-oriented understudy for gisting evaluation (ROUGE): ROUGE-1, -2 and -L. These were applied for every prediction to quantitatively assess LLM generation output quality. These metrics lie on a scale of 0.0 to 1.0, with the a higher value indicating better model performance. The BLEU metric is based on the modified $n$-gram (repeated terms), where *n* is a notation that set to between 1-4. In other words, BLEU measures local text patterns in a sentence by comparing the generated output to reference sentence(s), penalising deviations from the reference patterns [38].The more an LLM output deviates from this text, the lower the BLEU score. ROUGE-L is a recall-oriented metric that looks for the longest common subsequence between the reference and the candidate. ROUGE-1 and ROUGE-2 represent the specific number of n-grams [39]. The mean values and standard deviations of each metric were taken for each experiment, except for BLEU, which gives an overall score for the text corpus [38].

Metrics that assess generated text were found during the study not to always correspond to high-quality excipient recommendations; however, they did provide a useful baseline for assessing which LLMs maintained language proficiency. Consequently, a custom metric was designed to focus on the excipient accuracy called VELVET, which stands for Validation Excipient List Verification and Evaluation Tool. This was built to evaluate the similarity of a generated excipient to the actual excipient by comparing their embeddings. To evaluate the similarity between predicted and reference excipient compositions, we developed a functional embedding pipeline implemented in Python. The method embeds ingredient compositions into a continuous vector space and quantifies pairwise distances between predicted and reference formulations. Ingredient names were normalised using regular expressions to remove non-alphanumeric characters and convert text to lowercase. Redundant and template phrases (e.g., “for this formulation, use these excipients:”, and “this is printable and has … filament aspect”) were removed using case-insensitive regex substitution. Ingredient lists were tokenized by comma delimiters, and each entry was trimmed of whitespace. These preprocessing steps ensured consistent matching between predicted and reference ingredient names. The normalized ingredient–formulation matrix was transposed to form an ingredients × formulations matrix. Predicted and reference excipient compositions were extracted from a JSONL file containing paired “reference” and “prediction” text fields. Both fields underwent the same text normalization and ingredient parsing procedures as described above. For each prediction–reference pair, the Euclidean distance between the corresponding ingredient embeddings was computed using the cdist function from SciPy. Specifically, the mean pairwise distance between all predicted and reference ingredient embeddings was taken as the embedding distance metric. Instances with missing embeddings (i.e., ingredients absent from the embedding dictionary) were assigned NaN values, which were later replaced by the maximum observed pairwise embedding distance to penalize out-of-distribution predictions, which was 9.4475. Figure 2 is an illustrative representation of VELVET.

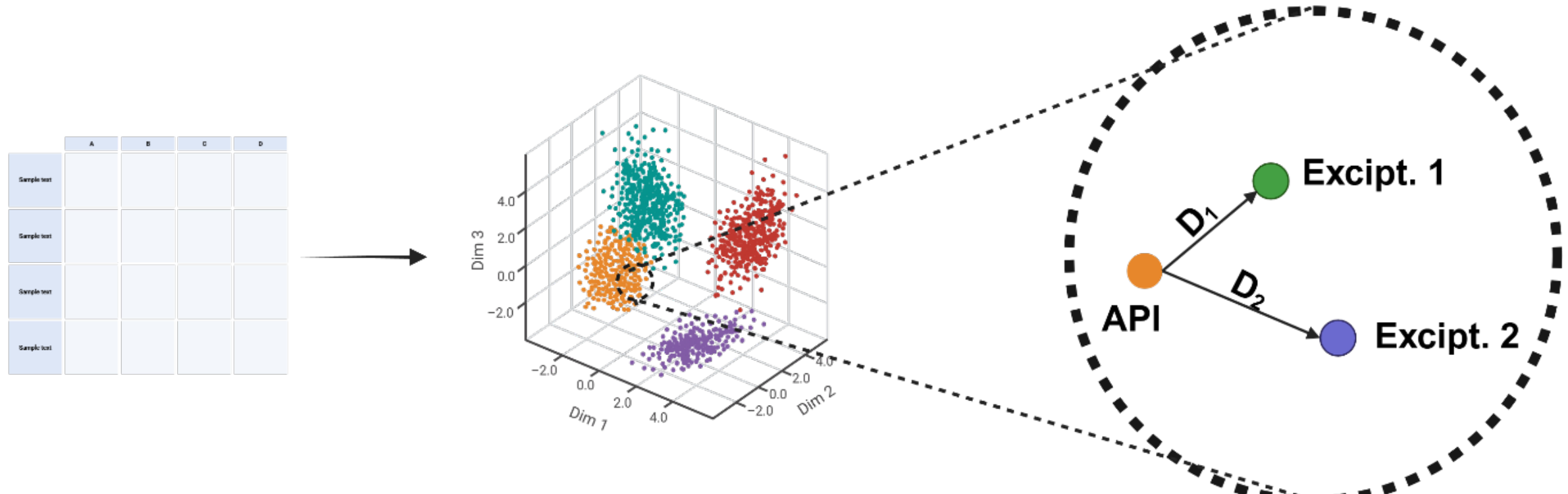

*Figure 2. Schematic depiction of VELVET. Tabulated formulation data was transformed into a high-dimensional embedding, where ingredient distances were determined based on the frequency of their co-occurrence. Points that are closer together in this embedding space represent ingredients that are more frequently used in combination. For VELVET, the average distance between the API and its predicted excipients was calculated, with lower VELVET scores indicating a higher likelihood that the excipients are used with the given API.*

Figure *3* portrays a schematic of the LLM pipeline.

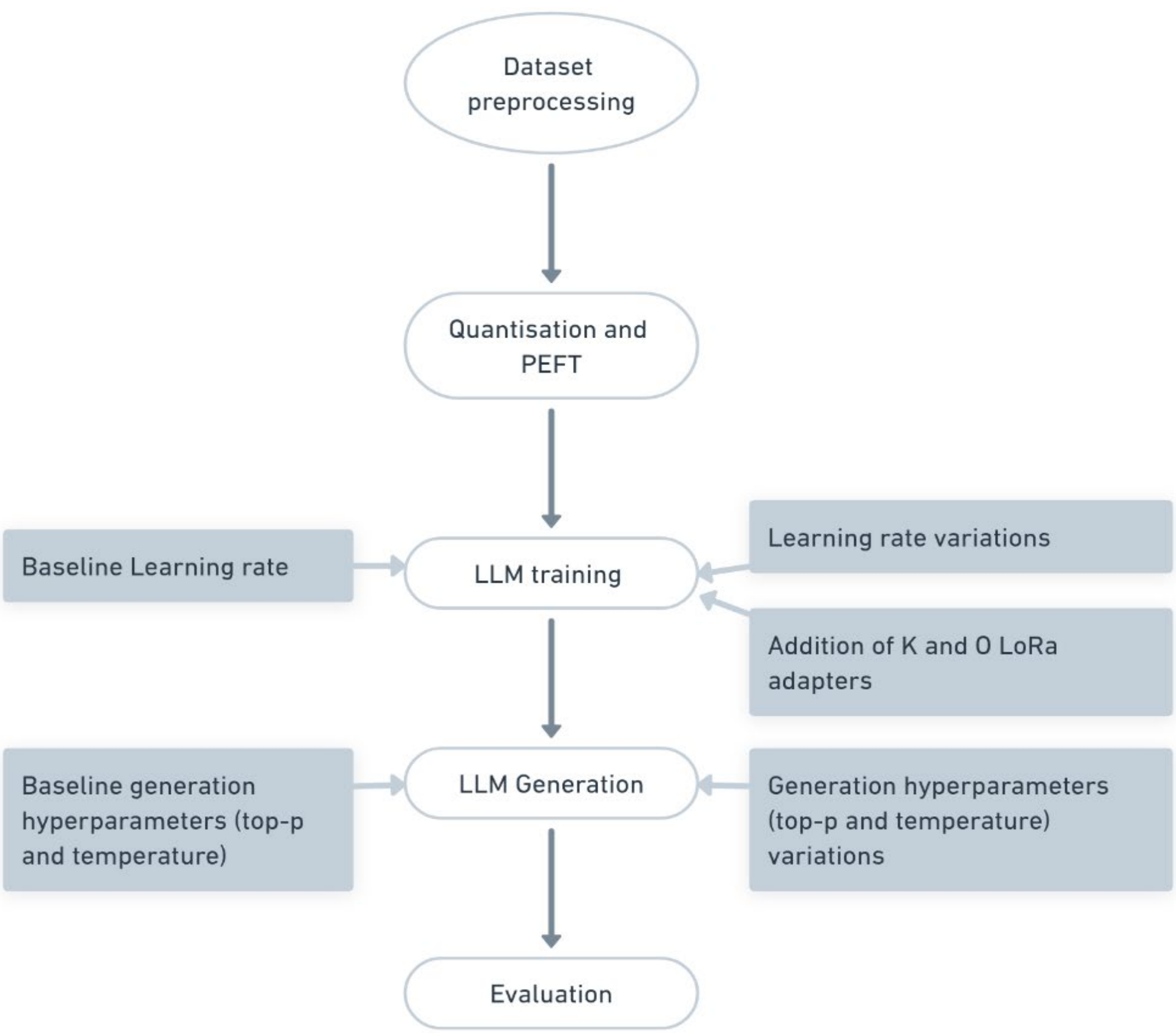

Figure 3. Schematic overview of the fine-tuning workflow applied to large language models (BioGPT, Mistral, Llama 2, and T5-XL) for excipient generation. The diagram illustrates the generalised flow of data preprocessing, Quantisation and PEFT. Then, the integration of LoRA-based parameter-efficient adaptation, learning rate and generation variations to the training and generation process. Then followed by an Evaluation of the performance of each LLM. Created using Whimsical.

# 3 Results

## 3.1 Dataset: Exploratory data analysis (EDA)

For PEFT of the LLMs, we used a large formulation dataset that was initially developed for traditional ML techniques. The dataset was structured in a vector-based structure, akin to a design of experiment (DoE) format but containing over 339 columns. This is because it contained over 61 APIs, over 276 excipients and the rest pertained to processing parameters. Figure 4 and Figure 5 present the most occurring APIs and excipients, respectively. Figure 6 presents the top 15 paired API-excipients. These analyses revealed that certain ingredients were used more often, and it will be interesting to see which ones are paired by the LLMs.

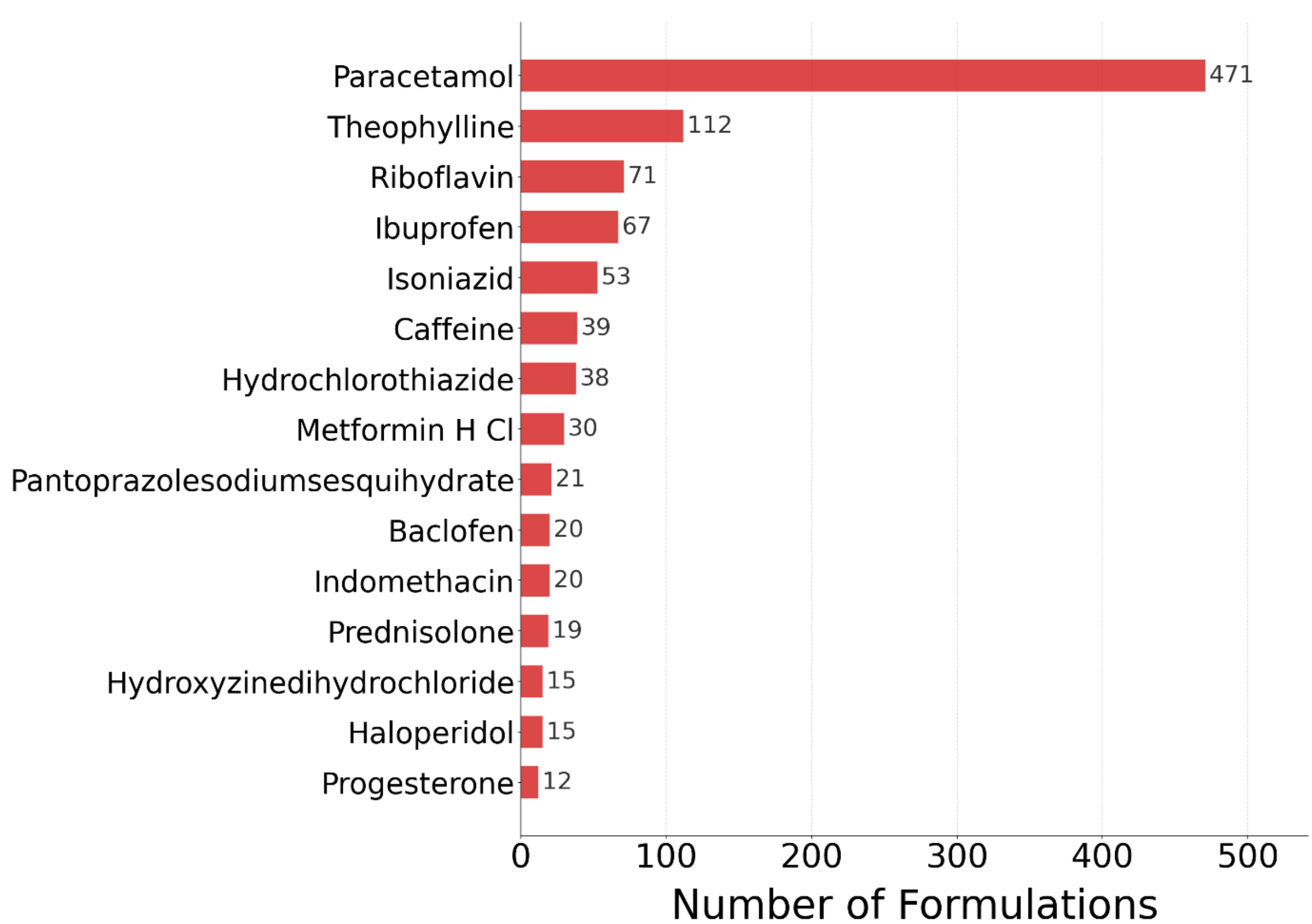

*Figure 4*. This figure illustrates the top 15 most frequent drugs in the dataset. The dataset's use of API is unbalanced and is weighted towards Paracetamol, which was used in 471 formulations, and Theophylline was used in 112. Therefore, Paracetamol is disproportionately mentioned to a much greater extent than all other drugs.

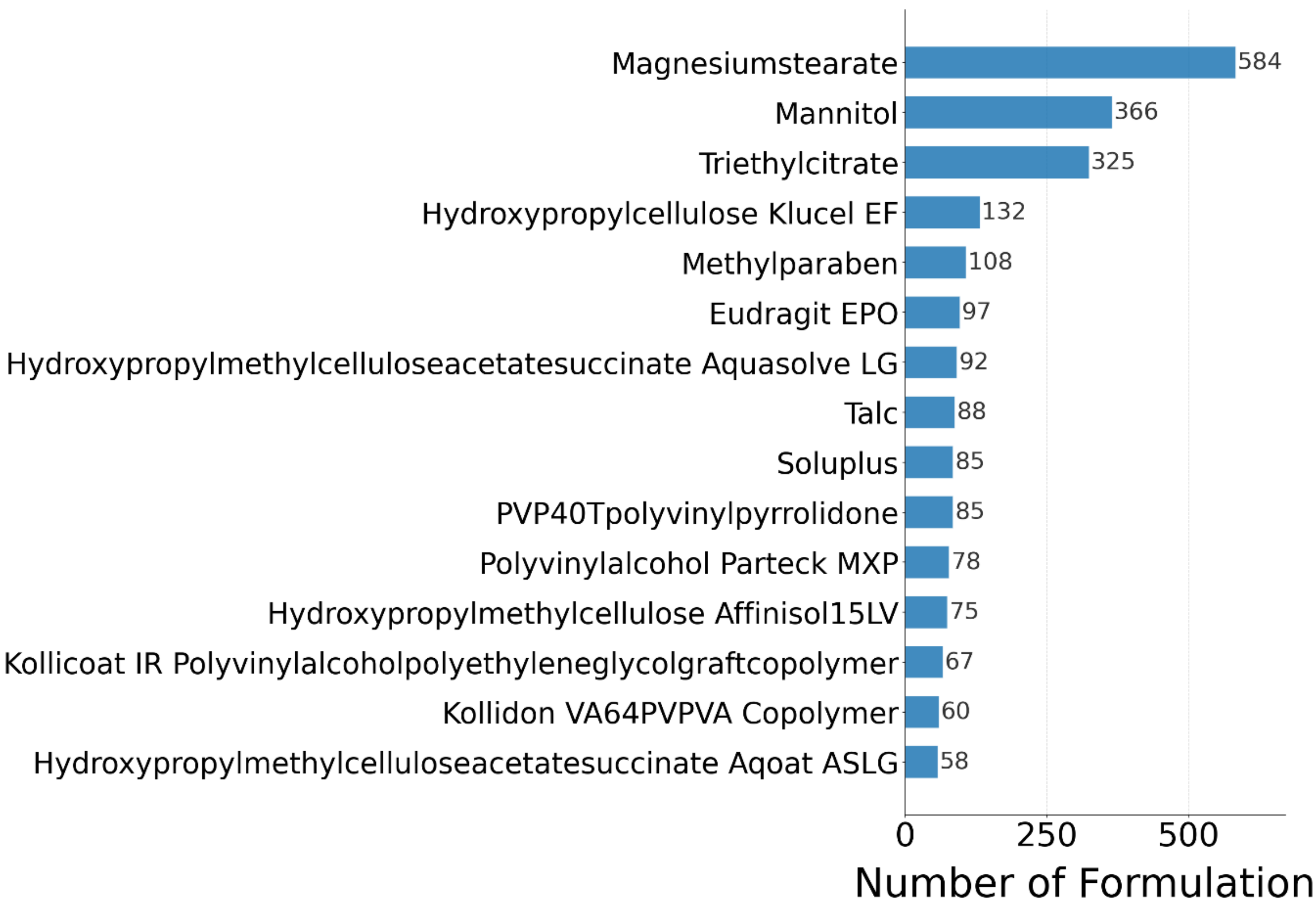

Figure 5. This figure illustrates the top *15* most frequent excipients in the drug dataset, *with*

*the top three* excipients *used being m*agnesium stearate *(584) mannitol (366) and* Triethyl citrate *(325).*

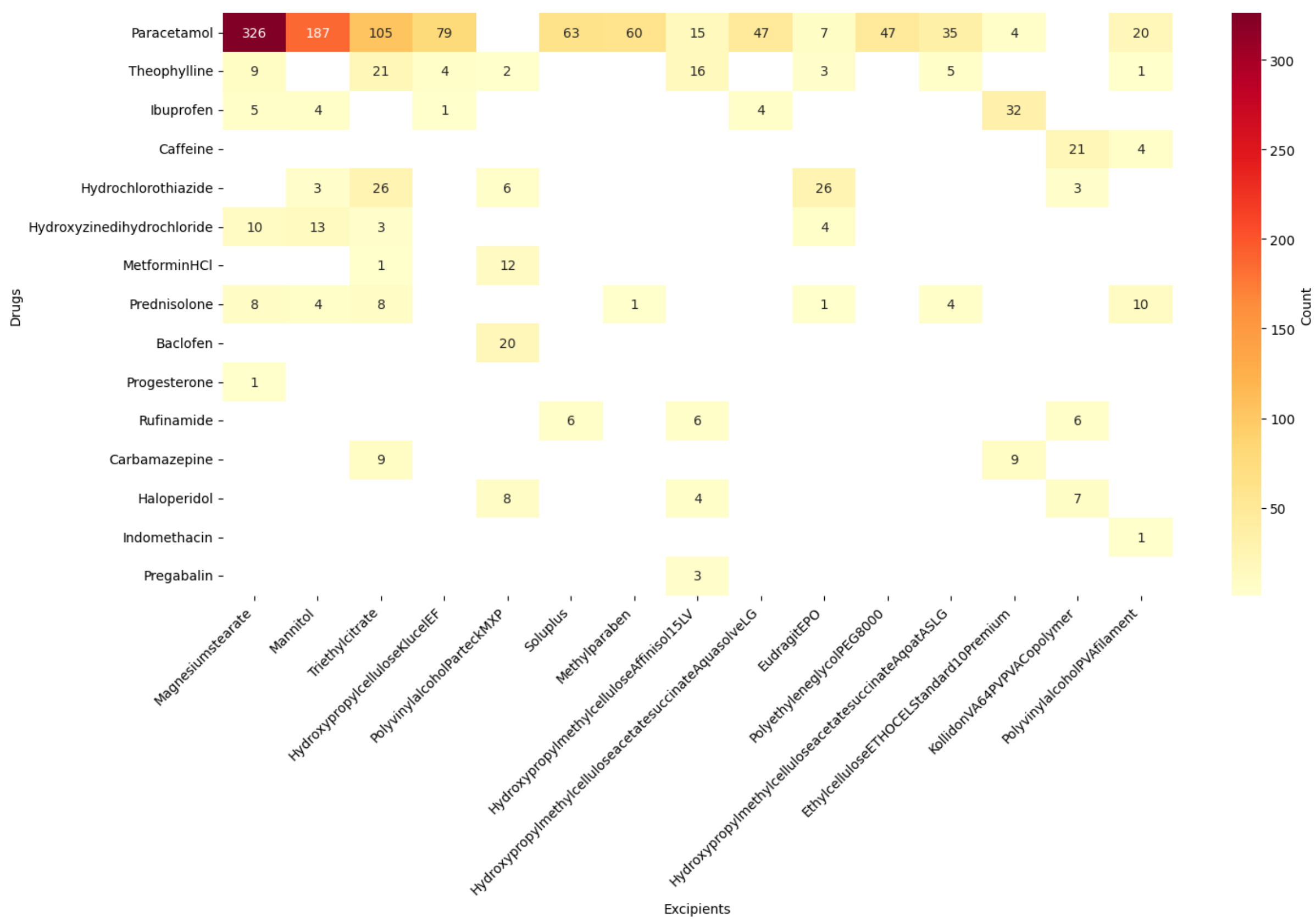

*Figure 6.* Heatmap depicting the most common drug-excipient combinations. This map highlights the most common excipients by count found within the APIs of the 3D formulation dataset.

Another key component of FDM is the mechanical properties of the filament, where filaments can be categorised as either good, flexible, brittle, unextrudable or unknown. The data was a combination of in-house and literature-acquired formulations, where not every filament mechanical properties were reported, hence the ‘unknown’ label. Unextrudable refers to formulations that were processed by a hot melt extruder, but did not form a viable filament. The remaining labels refer to filaments with optimal properties (good), filaments with high stiffness but susceptible to fracturing (brittle) and filaments that are susceptible to buckling (flexible). The ideal label is ‘good’ but considering that the process has been trial-and-error, a range of filament mechanical properties have emerged. The dataset contained a majority of filaments that were labelled as ‘Good’ (Figure 7).

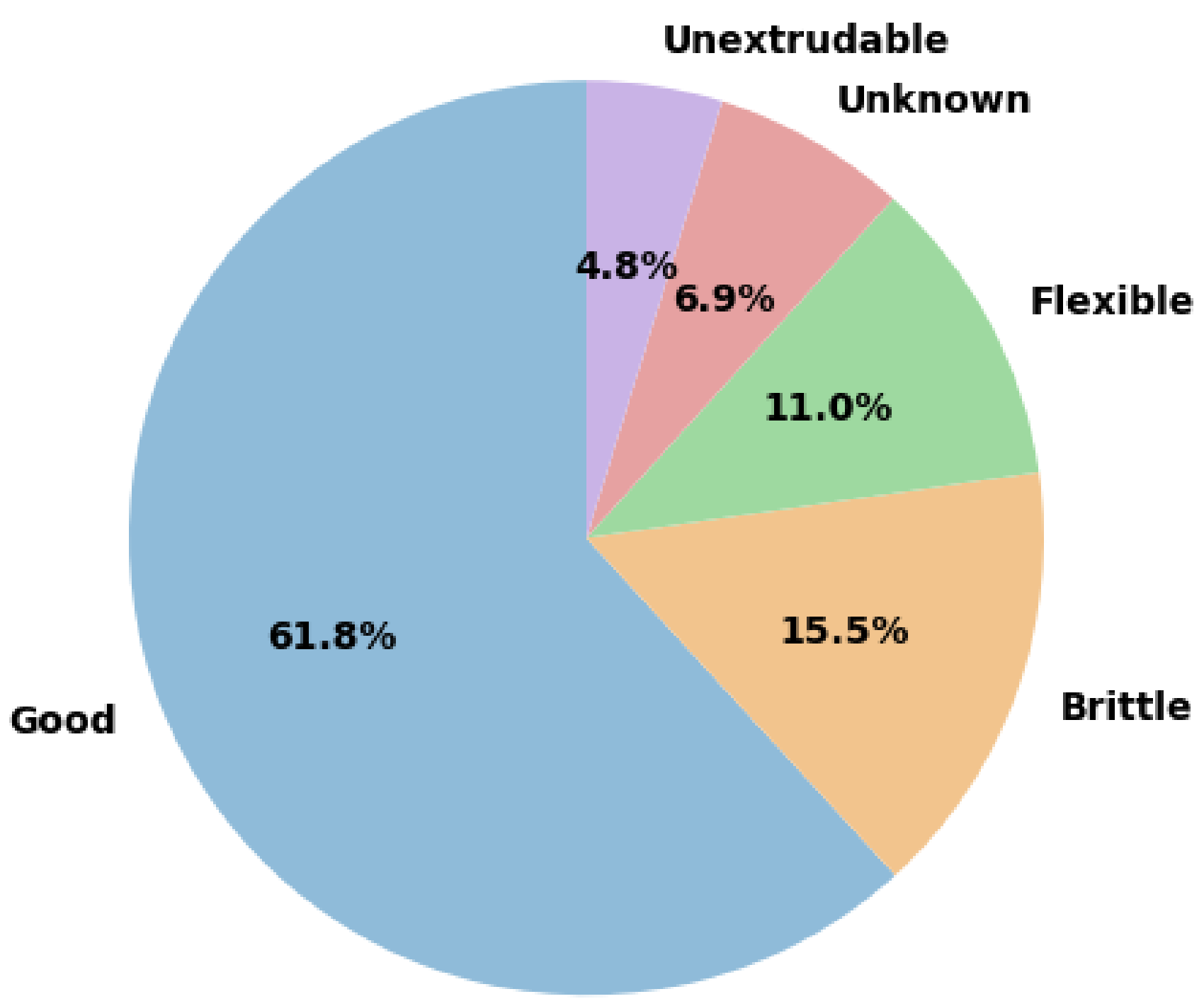

*Figure 7.* Pie chart shows the proportion of the formulation's filament aspect. The ideal training should provide a balanced dataset.

## 3.2 LLM Fine-Tuning Training

We experimented with four different LLMs, each given the same formulation dataset. These four exhibited distinct characteristics, and as it is the first experiment of its kind, we explored multiple LLMs. Once the data was fed to the LLMs, the loss curve was recorded during training. In a loss curve, it measures a model's error rate from the training data, with the lower the value after each training iteration (referred to as an epoch) should be decreasing until it converges. Herein, we limited the number of epochs to 4 because the training for the larger models was computationally expensive and time consuming. The training loss curve revealed that Llama 2 and Mistral had similar losses through the four epochs. By the fourth epoch, both Llama 2 and Mistral converged, with Llama 2 exhibiting a lower loss (0.1325) compared to Mistral (0.1830). T5-XL follows behind with a higher average loss across epochs, converging at a higher loss than Mistral and Llama 2 at 0.9339. BioGPT starts with a significantly higher loss at 5.6566 and converges at a much higher loss than the other LLMs at 2.9047 (Figure 8).

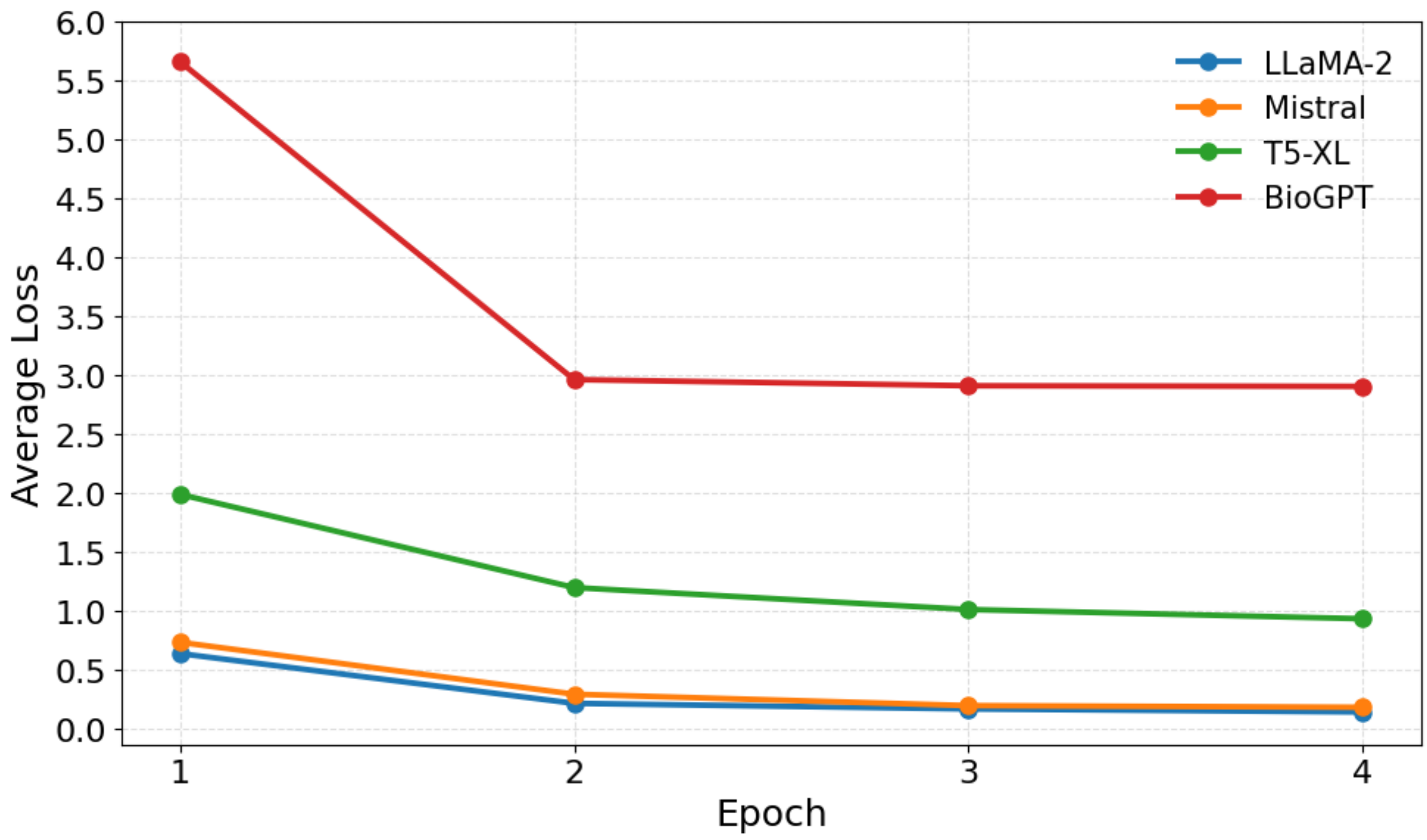

*Figure 8.* The average loss over epochs between each LLM. When the loss levels out into a straight line, it shows that the model has reached convergence, whereby the lowest loss has been achieved. The lower the loss, the more accurate the LLM's predictions were compared to the actual data. So, according to the average loss, Llama 2 has the best performance, followed closely by Mistral, then T5-XL. BioGPT, according to the loss, has the lowest predictive accuracy; it starts with a much higher loss and also converges well above the other LLMs.

## 3.3 Fine-tuning Evaluation

### 3.3.1 The Effect of Learning Rate

Following fine-tuning, we evaluated the LLMs using standard language metrics in BLEU and ROUGE. These were applied to evaluate the LLMs ability to generate comprehensive English outputs. For example, if prompted with: “Recommend me excipients to combine with 20 w/w% paracetamol”, should offer a linguistically meaningful response. Our Previous work found that learning rate is a key determinant in model performance and thus we explored the effect thereof on fine-tuned LLMs [13, 40]. Using the fastest learning rate of $10^{-2}$ results in faster model training times, however, LLM performance was poor, with BELU for most fine-tuned LLMs being close to 0.00 (Figure 9). For BioGPT, it could not be fine-tuned at this learning rate, which we speculate to be due to the developers’ hyperparameter selection.

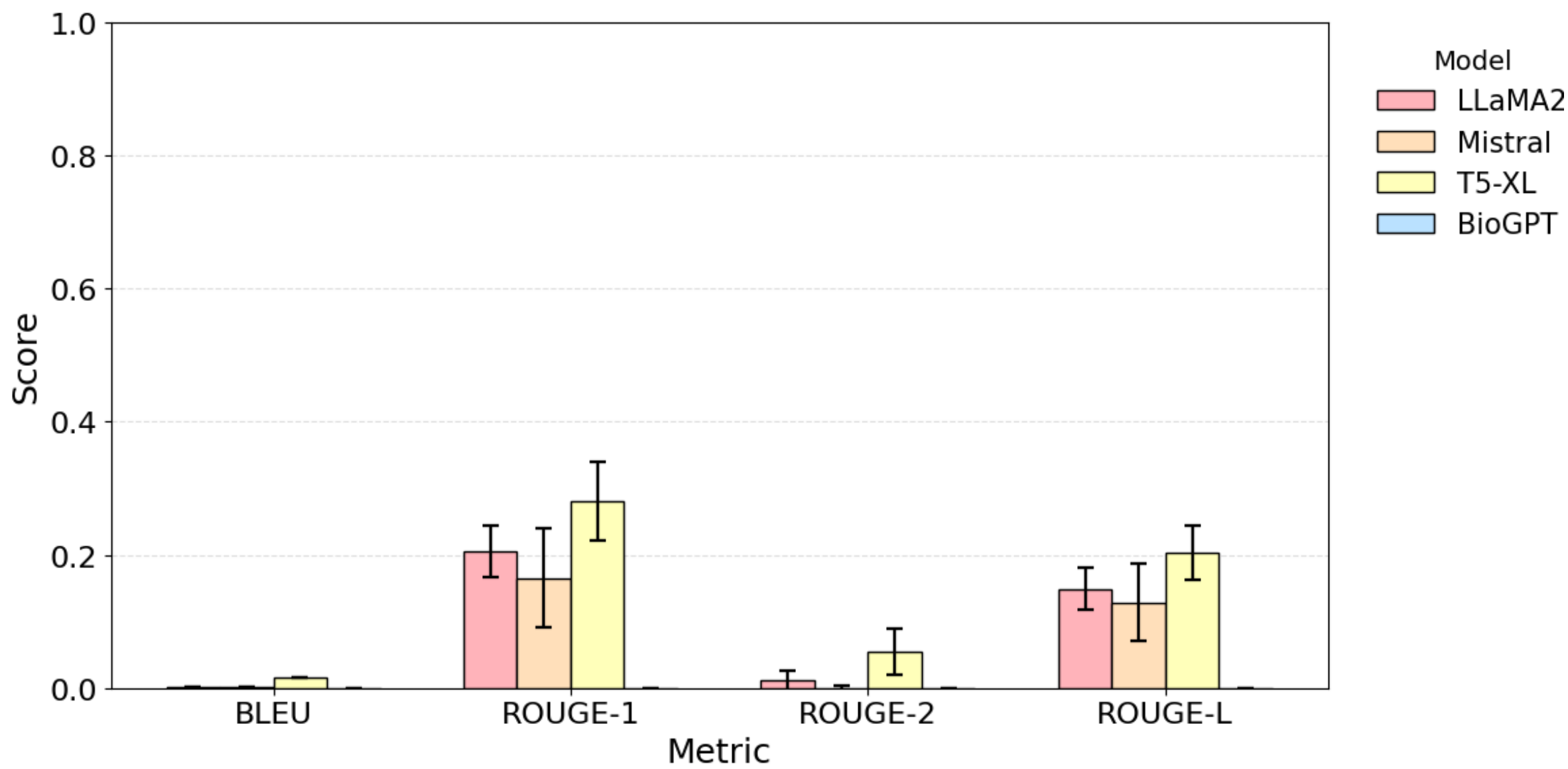

*Figure 9. The effect of $10^{-2}$ learning rate during fine tuning on LLM performance.*

Reducing the learning rate to $10^{-4}$ decreases the speed of fine tuning but it had a significant impact on model performance. Their BLEU and ROUGE scores were significantly higher, as observed in Figure 10. Llama2 and T5-XL were the better performers, producing a BLEU score of 0.62 and 0.53, respectively. Their ROUGE scores were also relatively high, where the ROUGE-1 was 0.78 ± 0.068 and 0.70 ± 0.12, respectively. In contrast, Mistral and BioGPT produced low metrics, where their BLEU scores were 0.26 and 0.24, respectively. Their ROUGE-1 score were also noticeably lower at 0.48 ± 0.14 and 0.47 ± 0.056. These values indicate that both Mistral and BioGPT struggled to produce coherent English outputs.

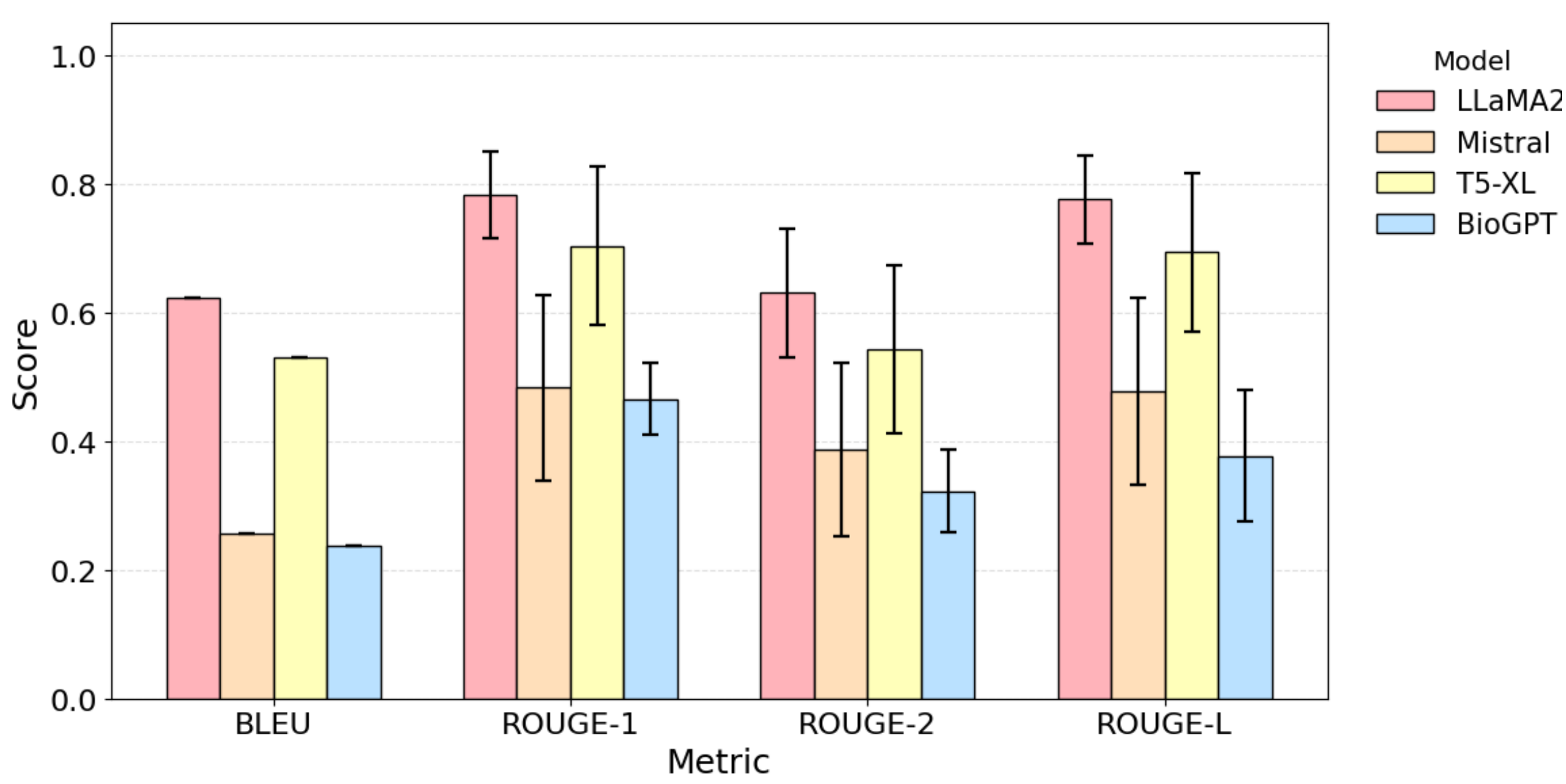

*Figure 10. The effect of $10^{-4}$ learning rate during fine tuning on LLM performance.*

The final learning rate investigated was $10^{-6}$, which required a longer period of time for training, on an already computationally demanding process. Here, a wide-spread decrease in model performance was observed (Figure 11) in comparison to $10^{-4}$. Furthermore, unlike $10^{-2}$, there were no issues in fine-tuning BioGPT. Overall, these findings are in agreement with previous work that indicate learning rate is a key determinant of model training, of which herein the optimal learning rate was $10^{-4}$. The findings also signalled that BioGPT may not be optimised for fine-tuning like the other LLMs.

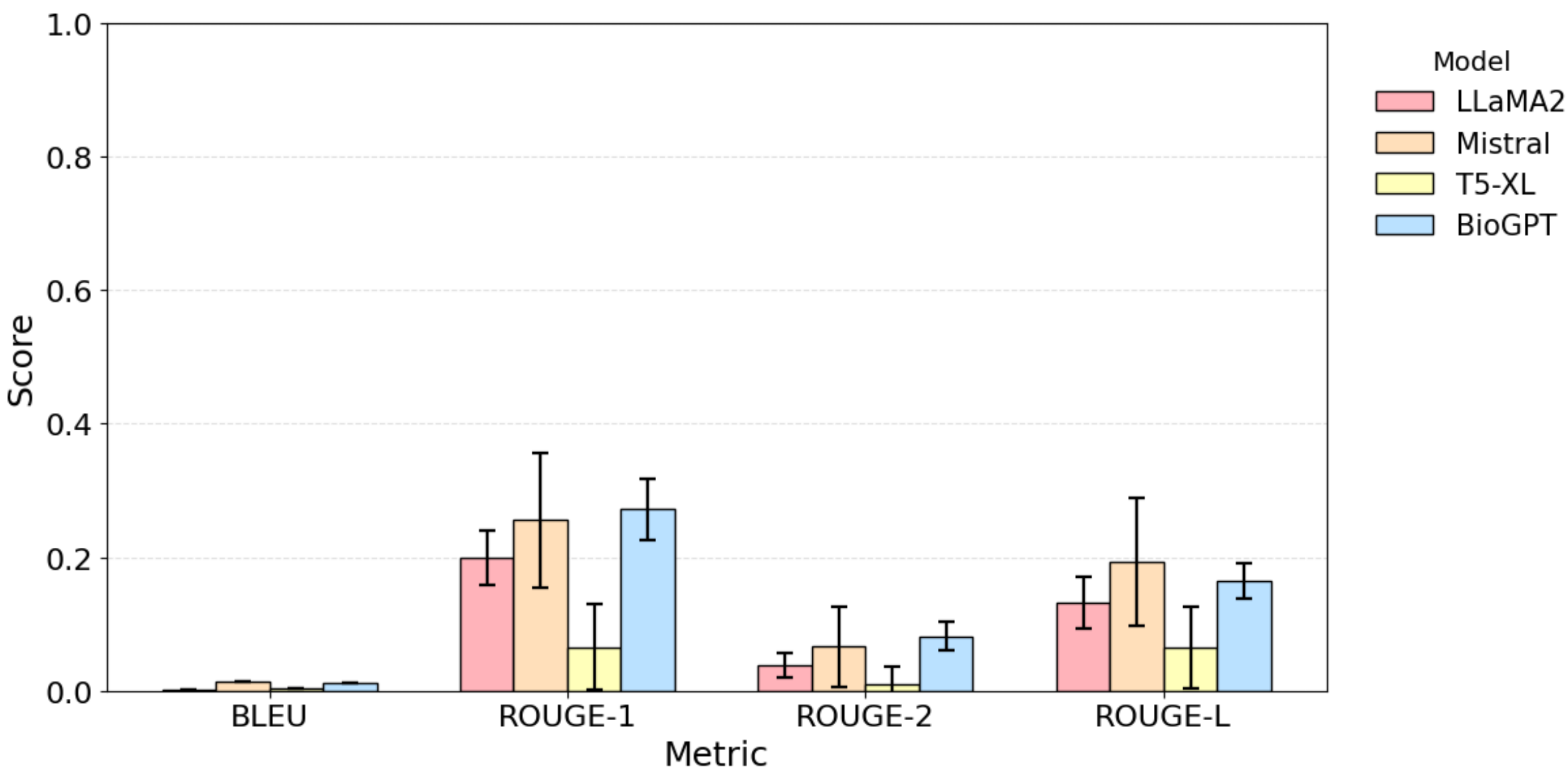

*Figure 11. The effect of $10^{-6}$ learning rate during fine tuning on LLM performance.*

We further inspected the outputs to gain a better understanding of LLM performance, and examples of outputs are presented in Table 4. Each LLM and their corresponding learning rate a text input of “Recommend excipients for 5 w/w% paracetamol”. For BioGPT, no learning rate resulted in a desirable output. While $10^{-2}$ failed to complete, $10^{-4}$ hallucinated excipient names (e.g., Polyvinyl Carbatesuccinatesuccinatesuccinate) and produced nonsensical alphanumeric text (i.e., \ub9dd\ucc45\uc5f4alternatinatinatinatinating). The learning rate of $10^{-6}$ changed the units from w/w% to w/v% and only recommended APIs. Mistral was also unable to produce comprehensible outputs. In contrast, Llama2 and T5-XL were able to produce comprehensible outputs at a learning rate of $10^{-4}$, insofar as producing real excipient names, reasonable combination of excipients (e.g., main polymer carrier, a plasticiser and a lubricant) and ensuring that the formulation totalled to 100 w/w%.

*Table 4. Example outputs generated by the fine-tuned LLMs across all three learning rates. The first row contains the text input and a reference example that was fed into the LLMs during the training process.*

| Input | Reference Output |
| --- | --- |

| Recommend excipients for 5 w/w% Paracetamol | For this formulation, use these excipients: Guar gum: 40 w/w%, Methyl paraben: 10 w/w%, Polyethylene glycol PEG8000: 5 w/w%, Shellac: 40 w/w%. This is printable and has a Good filament aspect. |
|---|---|
| **BioGPT output** | |
| Learning rate: $10^{-4}$ | "Given a drug and its quantity, suggest suitable excipients with their quantities, and indicate printability and filament aspect. Input: Paracetamol: for 5 w/w% Output: For this formulation, use these excipients: Polyvinyl Carbatesuccinatesuccinatesuccinate: 5 w/w%, Polyethyleneglycol1000: 5 w/w%, Polyethylene glycol stearate: 5 w/w%. This is printable and has a Good filament aspect. \ub9dd\ucc45\uc5f4alternatinatinatinatinating" |
| Learning rate: $10^{-2}$ | Learning rate was too high to produce an output “RuntimeError: CUDA error: device-side assert triggered” |
| Learning rate: $10^{-6}$ | Given a drug and its quantity, suggest suitable excipients with their quantities, and indicate printability and filament aspect. Input: Paracetamol: 5% w / v, paracetamol: 5% w / v, and aspirin: 2.5% w / v were selected as the drug model and excipients. |
| **Llama 2 output** | |
| Learning rate: $10^{-4}$ | For this formulation, use these excipients: Eudragit EPO: 45 w/w%, Magnesium stearate: 5 w/w%, Mannitol: 30 w/w%, Polyethylene glycol PEG8000: 15 w/w%. This is printable and has a Good filament aspect. |
| Learning rate: $10^{-2}$ | :\n and drug:: this a indicate:.0., aspect G#G. Input:#ientsability. : Response#.:G##,0##.\n\n 0##ability05.#ients and a acip0 use:0.##0 use:0 0 and indicate print.,0 and indicate and is,#ulation,:yl0##:,,For,,,# these fil,0 fil.0: and fil Mag,55.0.0yl aspect fil.0#5ament.1ament.000ol...0.0:0000010.00ethyl0able.5.000.2.00.00 fil0.0 print..0.0..::1...0:0.:.1..0000. ..0:0:0 |
| Learning rate: $10^{-6}$ | Excipient Quantity (mg) Printability Filament Aspect\nStarch 100 \u2611\ufe0f Printable \u2611\ufe0f\nMagnesium Stearate 5 \u2611\ufe0f Printable \u2611\ufe0f\nGlycerin 50 \u2611\ufe0f Printable \u2611\ufe0f\nSodium Lauryl Sulfate 2 \u2611\ufe0f Printable \u2611\ufe0f\n\nNote:\n\n\u2611\ufe0f: The excipient is suitable for the drug and can be used in the formulation.\n\n* Printability: The excipient can be printed with the drug using a 3D printer.\n* Filament aspect: The excipient can be used as a filament for 3D printing. |
| **Google T5-XL output** | |
| Learning rate: $10^{-4}$ | For this formulation, use these excipients: Hydroxypropyl cellulose Klucel EF: 45 w/w%, Magnesium stearate: 5 w/w%, Mannitol: 15 w/w%. This is printable and has a Brittle filament aspect. |
| Learning rate: $10^{-2}$ | "input": "Paracetamol: 5 w/w%", "reference": "For this formulation, use these excipients: Guar gum: 35 w/w%, Hydroxypropyl methylcellulose acetate succinate Aquasolve LG: 35 w/w%, Methylparaben: 5 w/w%, Polyethylene glycol PEG8000: 5 w/w%, Talc: 15 w/w%. This is printable and has a Good filament aspect.", "prediction": ". this This. this:: this, this formulation<br>: excip ex exs use ex use this thesecip these,:cip.: Thiscip ex, this has these these this this ex.cip iscip this these excip ex these: formulation this this uses this these this ex" |
| Learning rate: $10^{-6}$ | "input": "Paracetamol: 5 w/w%", "reference": "For this formulation, use these excipients: Guar gum: 45 w/w%, Polycaprolactone MW80000: 45 w/w%, Polyethylene glycol PEG8000: 5 w/w%. This is printable and has a Good filament aspect.", "prediction": "glycerin: 25 w/w%" |
| **Mistral output** | |
| Learning rate: $10^{-4}$ | For this formulation, use these excipients: Eudragit L100: 45 w/w%, Magnesium stearate: 5 w/w%, Mannitol: 20 w/w%, Methyl paraben: 15 w/w%, Triethyl citrate: 10 w/w%. This is printable and has a Brittle filament aspect.\n\n### Explanation:\nThis is a solid formulation which can be directly printed. |
| Learning rate: $10^{-2}$ | .\n,. :. .. ### exc aspect suggest.0ip:\nable a######,..0\nable5:\n and0\n,0:\n aspectiven:, |
| Learning rate: $10^{-6}$ | ```\nExcipients:\n1. Polyvinyl Alcohol: 0.15\n2. Talc: 0.02\n3. Triethyl Citrate: 0.01\n4. Magnesium Stearate: 0.01\n\nPrintability: Good\nFilament Aspect: Smooth\n```\n\n### Explanation:\nThe given drug is Paracetamol with a quantity of 0.05. Based on the drug and its quantity, the suggested excipients with their quantities are Polyvinyl Alcohol: 0.15, Talc: 0.02, Triethyl Citrate: 0.01, and Magnesium Stearate: 0.01. The printability is good and the filament aspect is smooth. This means that the suggested formulation can be easily printed using a 3D printer |

## 3.3.2 The Effect of LoRA Adapters

We next explored the effect of LoRA adapters, which are known to impact model performance, which are low-rank matrices (Figure 1) integrated into the model's attention and projection layers to enable efficient fine-tuning without modifying the original model weights [41]. Specifically, we looked at the 'K' and 'O' adapters, where 'K' corresponds to 'key' and 'O' refers to 'output' projection adaptation. In essence, they are additional parameters that can be adjusted to affect model performance without needing to fully train an LLM. Keeping the learning rate at the optimal value of $10^{-4}$, these additional parameter tuning resulted in marked improvements for both Mistral and marginal improvements for BioGPT (Figure 12). Mistral's generative capacity improved after the addition of those LoRA adapters, with the BLEU scores increased from 0.26 to 0.51, and the ROUGE-1 (0.48 ± 0.14 to 0.66 ± 0.14), ROUGE-2 (0.39 ± 0.14 to 0.51 ± 0.14) and ROUGE-L (0.48 ± 0.15 to 0.64 ± 0.15), indicating broad generative improvement. A t-test comparing the before and after the addition of LoRA adapters confirmed a statistically significant increase in Mistral but not for BioGPT, Llama2 nor T5-XL. Table 5 presents the responses of BioGPT after the addition of these parameters, where responses were improved, albeit containing alphanumeric text in BioGPT's outputs.

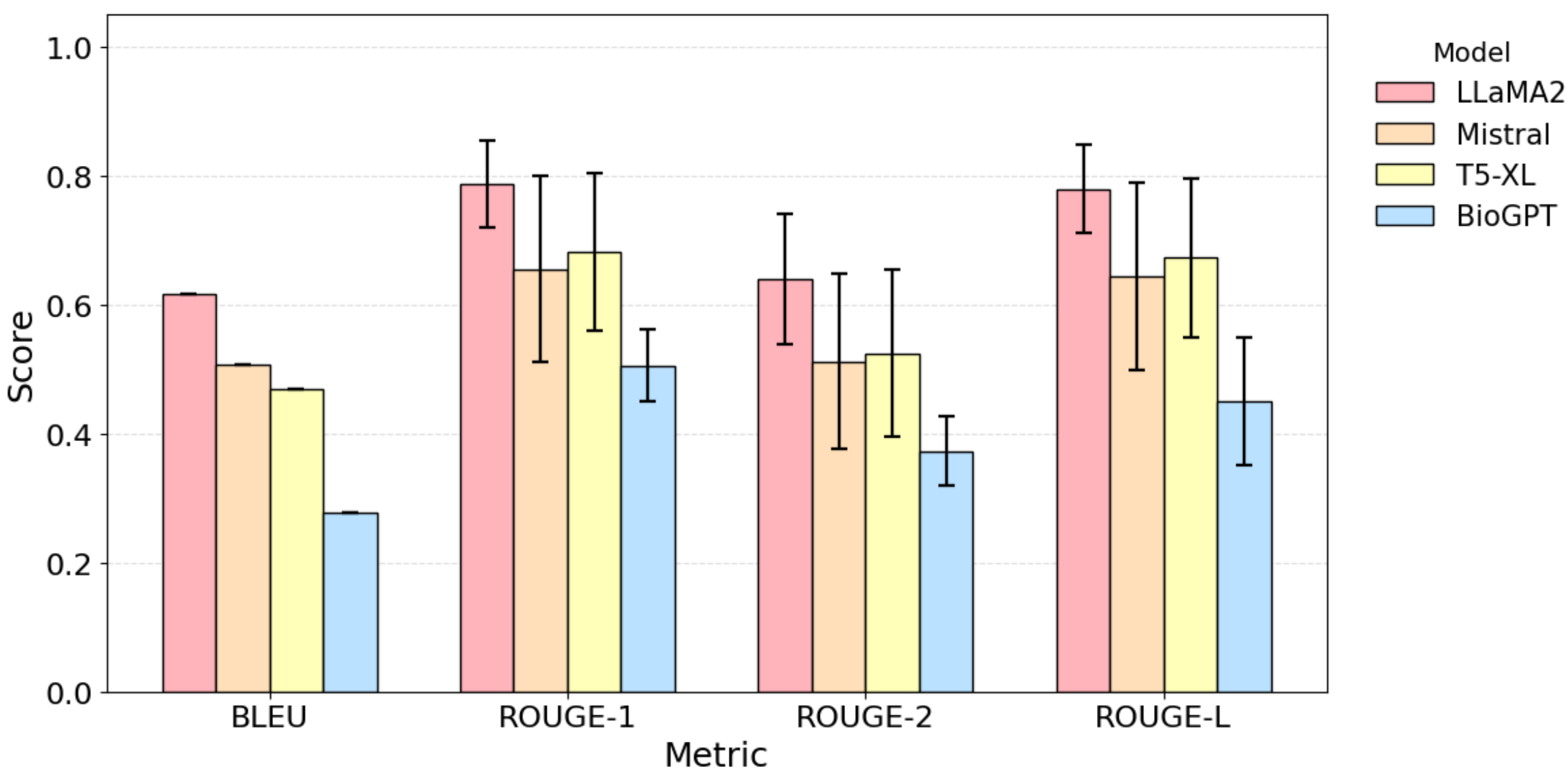

*Figure 12. Comparative bar plot shows the effect that the addition of 'K' and the 'O' LoRA adapters have on the LLMs. Demonstrated by the metric values of each one. Notably Mistral and BioGPT's metrics showed improvement.*

*Table 5. The effect of additional projection layers 'K' and 'O' LoRA adaptors on generative performance for BioGPT. Ideally, the predictions should closely resemble the reference.*

| **Input** | **Reference Output** |
|---|---|
| Recommend excipients for 5 w/w% Paracetamol | For this formulation, use these excipients: Guar gum: 40 w/w%, Methyl paraben: 10 w/w%, Polyethylene glycol PEG8000: 5 w/w%, Shellac: 40 w/w%. This is printable and has a Good filament aspect. |
| | **BioGPT output** |
| Before | For this formulation, use these excipients: Polyvinyl Carbate succinatesuccinatesuccinate: 5 w/w%, Polyethyleneglycol 1000: 5 w/w%, Polyethylene glycol stearate: 5 w/w%. This is printable and has a Good filament aspect. \ub9dd\ucc45\uc5f4alternatinatinatinatinating" |
| After | For this formulation, use these excipients: Polyvinylalcohol PVA PVA filament: 15 w/w%, Magnesium stearate: 5 w/w%, Mannitol: 15 w/w%. This is printable and has a Good filament aspect. \u0434, and.. \uc904 \u019aof a imatinatin |
| **Input** | **Reference Output** |
| Recommend excipients for 10 w/w% Clobetasol Propionate | For this formulation, use these excipients: PolylacticacidPLAIngeo4043D: 30 w/w%, Polyvinyl alcohol MOWIFLEXC 17 Mn 66000: 60 w/w%. This is printable and has a Good filament aspect. |
| | **BioGPT output** |
| Before | For this formulation, use these excipients: Polyethyleneglycol--PE: 20 w/w%, Polyethylene glycol- Polyethyleneglycolstearate: 30 w/w%, Magnesium stearate: 5 w/w%, Magnesium stearate: 10 w/w%. This is printable and has a Good filament aspect. \u2468 \uad6d polyethylenedic |
| After | For this formulation, use these excipients: Hydroxypropyl cellulose acetate succinate AqoatAqAW: 40 w/w%, Magnesium stearate: 10 |

| | w/w%, Magnesium stearate: 10 w/w%. This is printable and has a Good filament aspect. \ub9dd\uad6d |
|---|---|
| **Input** | **Reference Output** |
| Recommend excipients for 10 w/w% Amlodipine besylate | For this formulation, use these excipients: Polyvinyl alcohol Parteck MXP: 63 w/w%, Sorbitol: 27 w/w%. This is printable and has a Good filament aspect. |
| **BioGPT output** | |
| Before | For this formulation, use these excipients: PolyhydroxyethylcelluloseKluctratekkateKluctritoseKluctritkOnicitose: 40 w/w%, PolyhydroxyethylcelluloseKluctritkKluctritoMagnesiumstearate: 40 w/w%, Poly |
| After | For this formulation, use these excipients: Eudragit L100: 50 w/w%, Polyethylene glycol PEG8000: 5 w/w%, Polyethylene glycol PEG6000: 10 w/w%. This is printable and has a Good filament aspect. \u044f\u3048 \u0434 \uc8fdatinatinatinatinatinatinatin |

## 3.3.3 Domain-Relevant Metric

When inspecting the outputs, we realised that the language metrics (i.e., BLEU and ROUGE) were not suitable for determining the quality of the excipients recommended. In some outputs, the LLM will output unconventional mixture that, for example, that excluded a key ingredient like a lubricant (Figure 5). For that reason, we created a new metric called VELVET, which stands for validation excipient list verification and evaluation tool. It essentially compares the frequency of an excipient to API in an embedding space, where the closer the excipient is to an API the more it is used. Since it is a distance-based metric, the lower the VELVET score the better the model's performance. As expected, the learning rate of $10^{-4}$, and for some LLMs with additional adapters, yielded the lowest and better VELVET scores. Llama 2 produced the lowest VELVET score at a learning rate of $10^{-4}$ with no additional adapters, which was 4.03 ± 1.40, which when inspecting the recommended formulation across different inputs, corresponded with a good degree of similarity between excipients in the reference and generation. BioGPT produced the second lowest VELVET score at 4.7 ± 2.95. On this note, it is worth highlighting that VELVET does not take into consideration the linguistic performance, and only analysis the outputted excipients. Hence, it is why BioGPT can output alphanumeric text with additional LoRA adapters (Table 5), but can still generate a relatively low VELVET score. This suggests that BioGPT could potentially lead to high-performing LLM as a tool to accelerate pharmaceutical 3D printing formulations if the issue with the language output can be addressed.

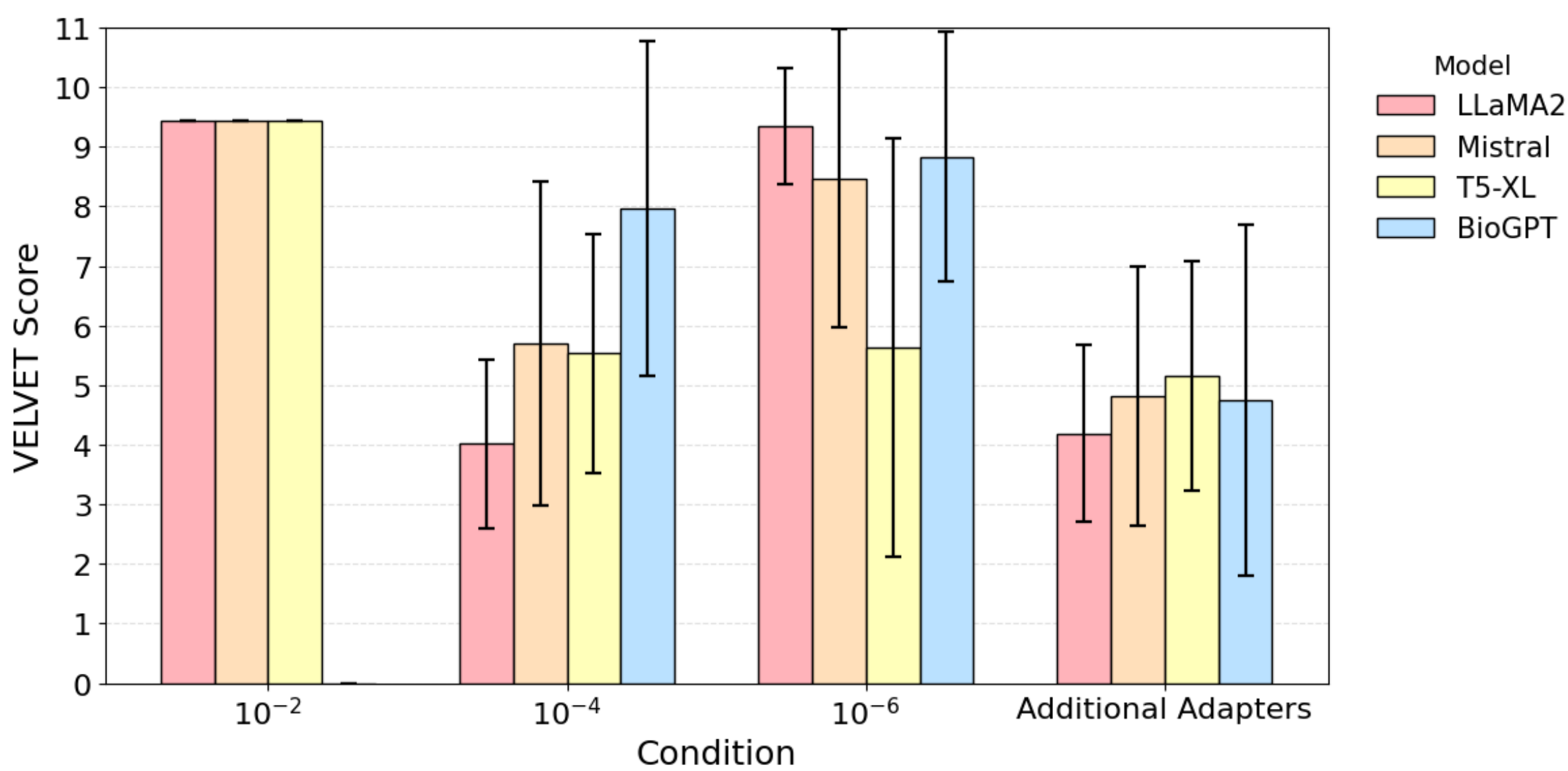

*Figure 13. The effect of learning rate and the addition of LoRA adapters on VELVET score. Note that the additional adapters were performed on the optimal learning rate of $10^{-4}$.*

## 3.4 The effect of Generative Parameters

Following fine-tuning, there are a set of parameters that can be optimised/controlled to boost model capabilities. Here we explored the two most common, which are 'temperature' and the nucleus sampling referred to as 'top p'. These allow a trained LLM to control the randomness and diversity of its generated outputs during inference. Adjusting these parameters had no effect on Llama2 across all metrics (Figure 14 to Figure 18), indicating that the generative parameters had less of an impact than the PEFT-related parameters. For T5-XL, changing the temperature value to 0.3 or 0.5 considerably reduced the BLEU score, from approximately 0.50 to 0.12. For Mistral, adjusting either the top p or the temperature from their default setting reduced its BLEU score. In contrast, adjusting the generative parameters yielded improvements in BLEU for BioGPT.

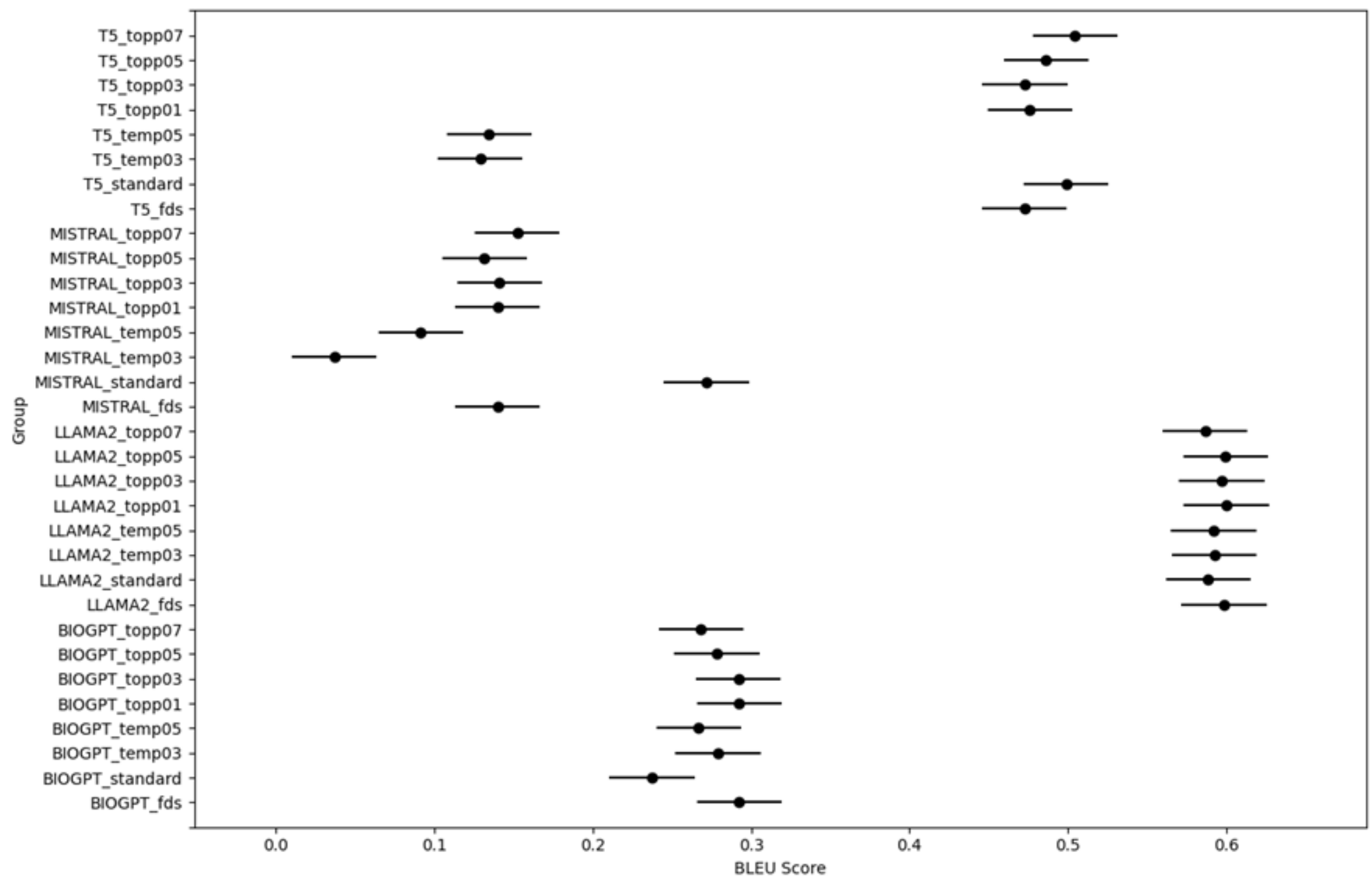

*Figure 14. The effect of LLM temperature and top p on BLEU scores. The 'fds' refers to setting the due sampling to 'false', which sets both parameters to 0. The standard are when the temperature and top p are set to 0.7 and 0.9, respectively.*

The same impact was observed for the three ROUGE metrics, where the same generative parameters that impacted an LLM's BLEU score also impacted their ROUGE. T5-XL generative performance shows more variation across temperatures, as T5-XL at a temperature of 0.5 appears to have a negative impact on ROUGE-L compared to the other temperatures. Mistral overall struggles with generative quality, but benefits from a higher temperature; for example, the ROUGE-L goes from 0.292 at 0 temperature to 0.4777 at 0.7.

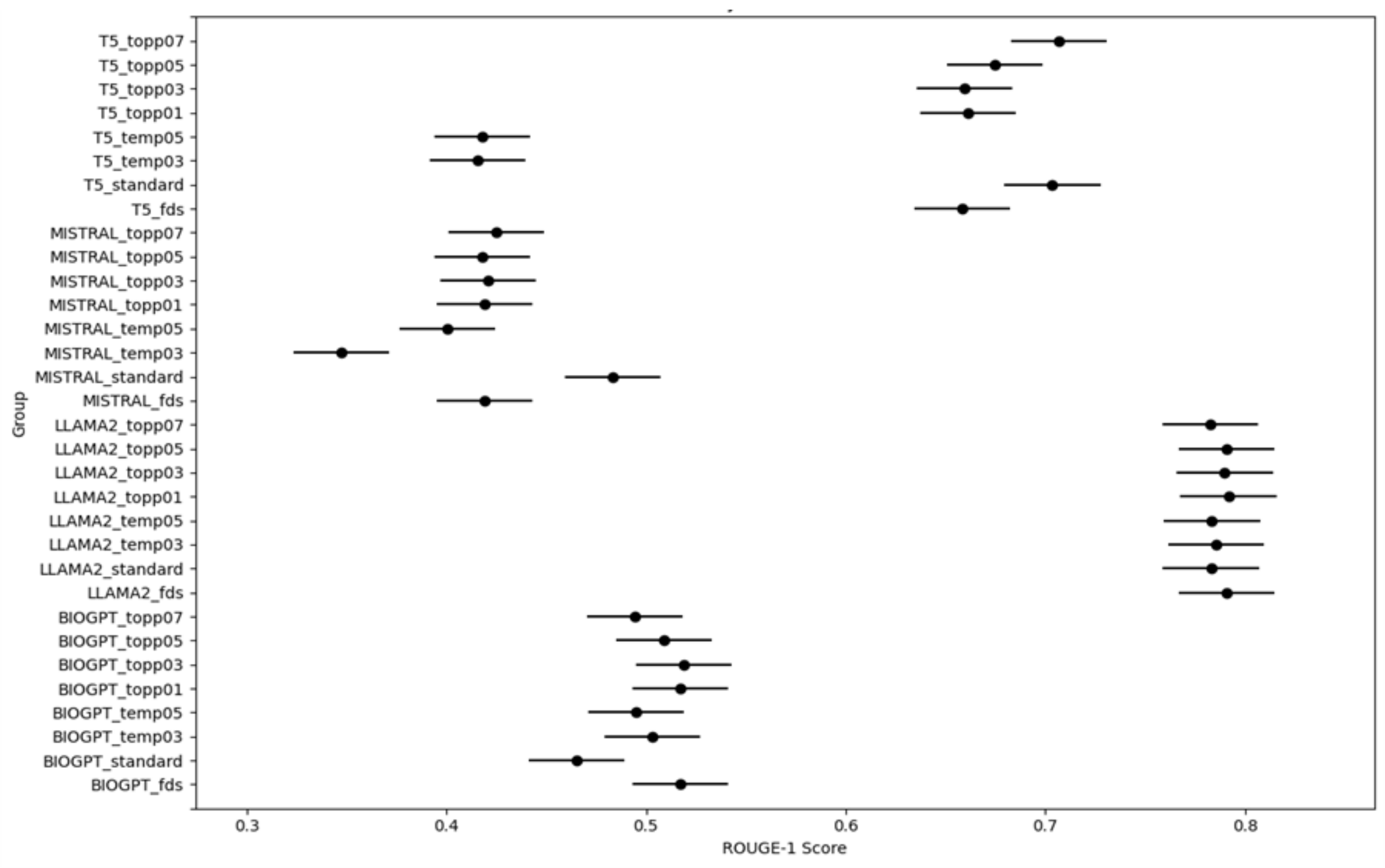

*Figure 15. The effect of LLM temperature and top p on ROUGE-1 scores.*

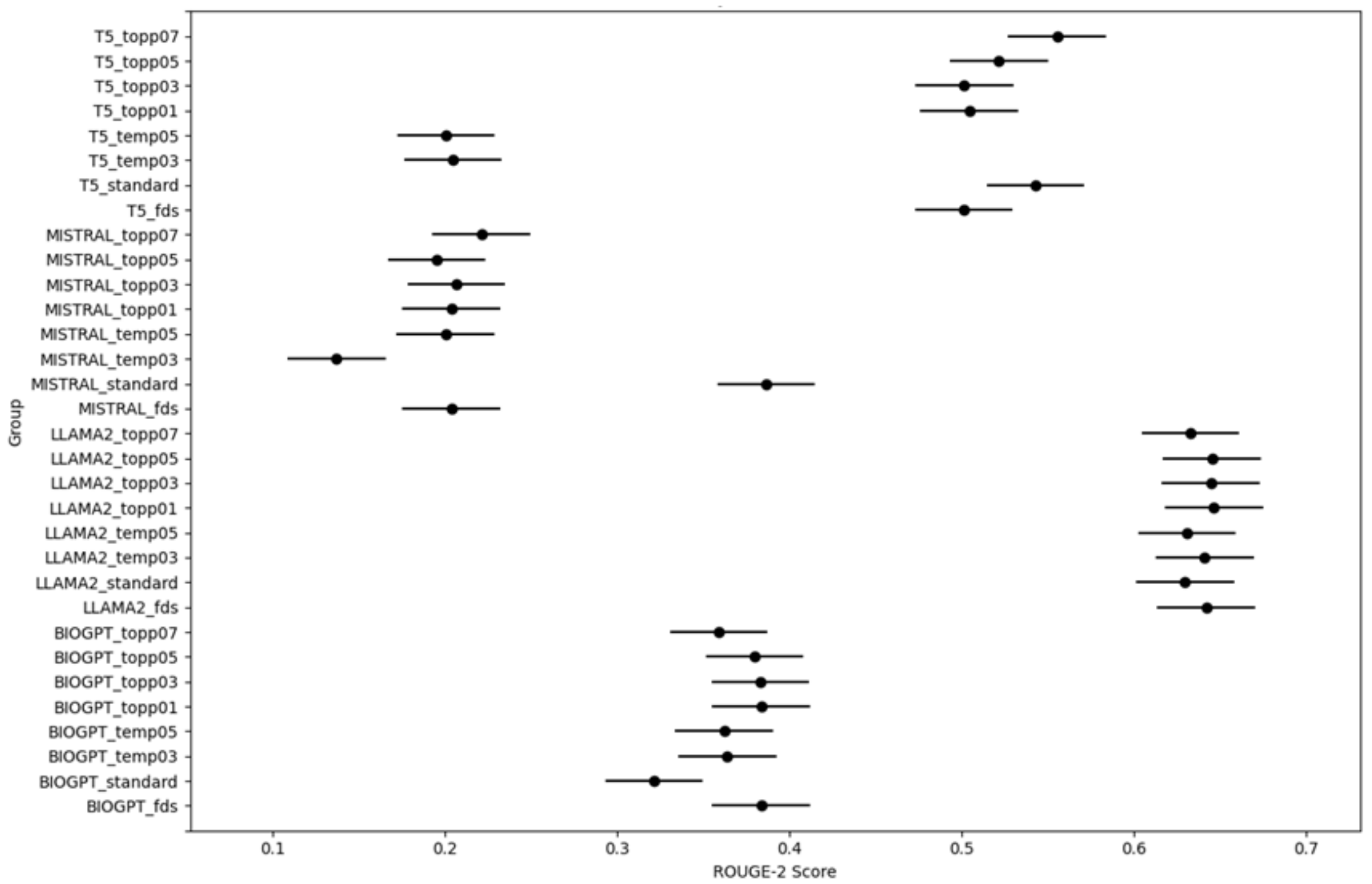

*Figure 16. The effect of LLM temperature and top p on ROUGE-2 scores.*

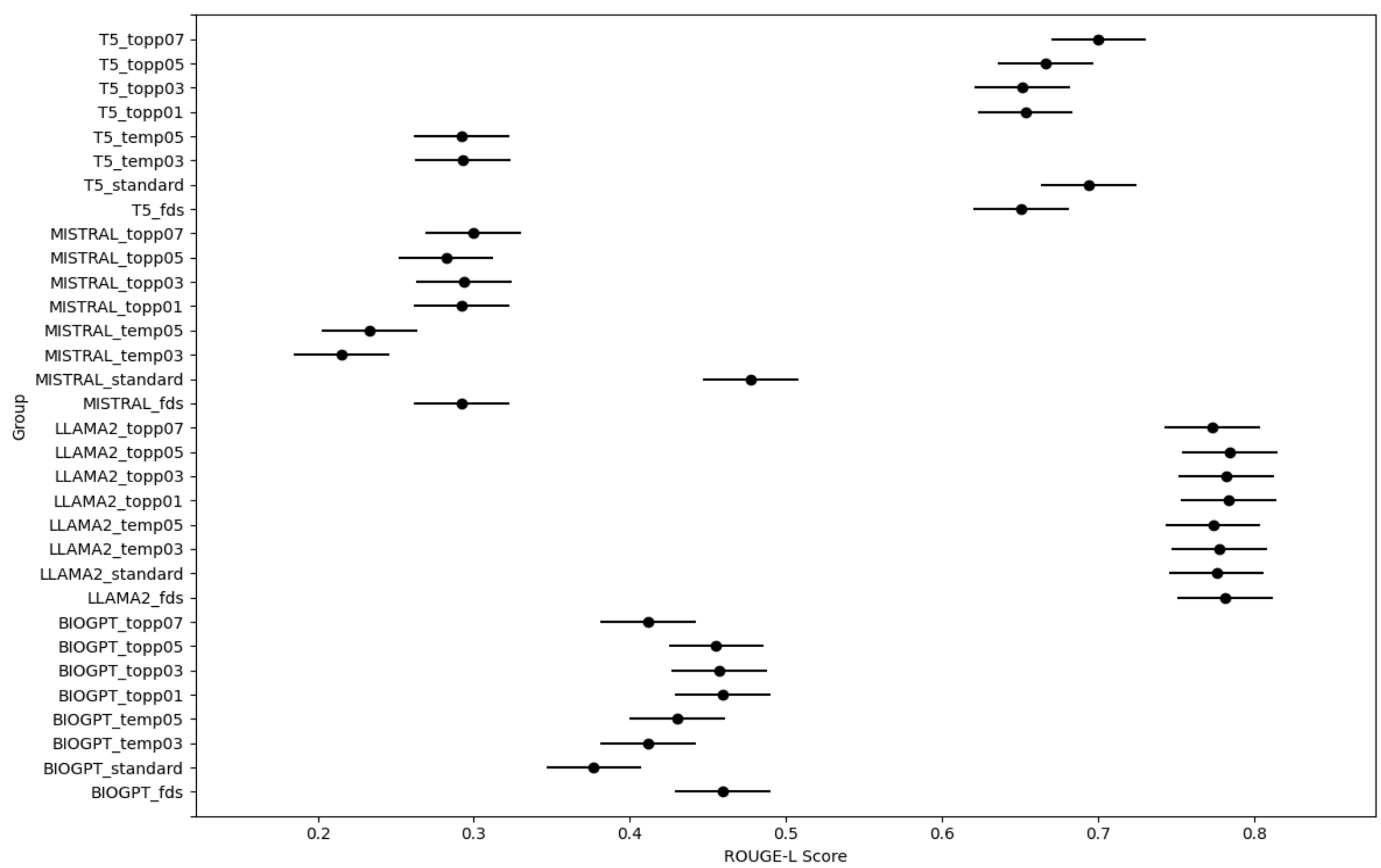

*Figure 17. The effect of LLM temperature and top p on ROUGE-L scores.*

The same was also observed for the VELVET metric, with Llama 2 producing the best scores (i.e., lowest), demonstrating minimal effect with changing the generative parameters, whereas the other LLMs were susceptible to change. What's interesting was that BioGPT improved on its default performance with changing of the parameters, particularly the top p values.

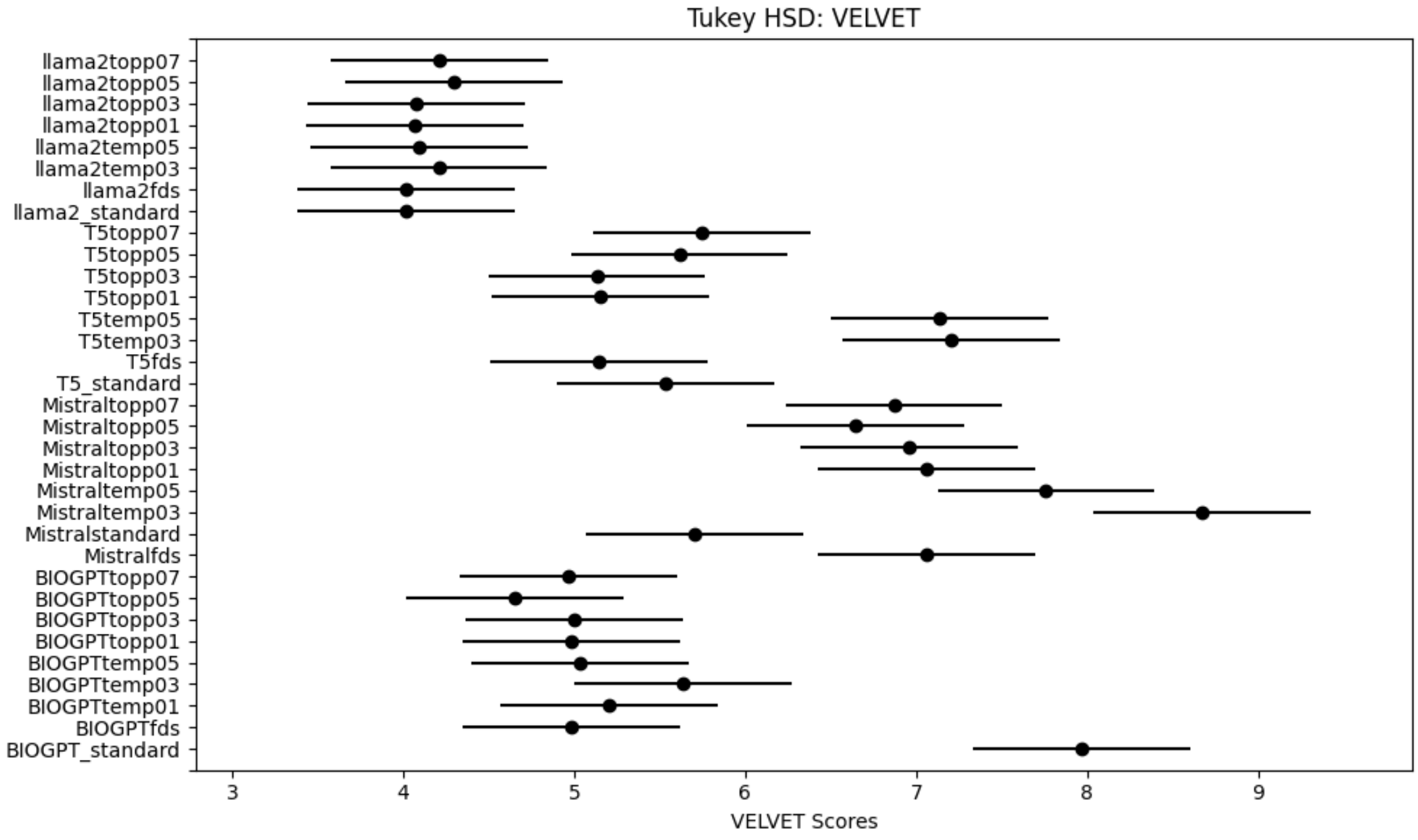

*Figure 18. The effect of LLM temperature and top p on VELVET scores.*

Overall, investigating the effect of generative parameters revealed that it can impact the performance of 3 out of the 4 LLMs. Aside from Llama2, it indicated that LLMs were sensitive to generative parameters, of which their degree of sensitivity varied.

## 3.5 Computational Evaluation

At the optimal learning rate of $10^{-4}$, the training times ranged from 3 minutes 42 seconds to 16 minutes 32 seconds (Table 6). These training times reflect the LLMs' parameter size, where the larger the parameter size the more time needed for training but not necessarily the number of parameters trained using LoRA. These models were trained using GPUs, which allowed them to be on the same time scale as traditional ML approaches [12].

*Table 6.* The training time for fine-tuning each LLM compared to its parameter size.

| LLM | Training time (minutes and seconds) | Total number of parameters | Number of parameters trained with LoRA |
|---|---|---|---|
| Llama 2 | 14:28 | 7,000,000,000 | 8,388,608 |
| Mistral | 16:32 | 7,248,547,840 | 6,815,744 |
| Google-FLAN-XL | 5:32 | 2,859,194,368 | 9,437,184 |
| BIOGPT | 3:42 | 349,908,992 | 3,145,728 |

# 4 Discussion

During the evaluation of multiple LLMs, we identified the need for task-specific evaluation metrics tailored to pharmaceutical formulation use cases. As the ecosystem of LLMs, and their corresponding small language model (SLM) variants [42], continues to expand, reliance on metrics focused solely on linguistic competence is insufficient for pharmaceutical applications. Instead, additional evaluation criteria are required to assess a model's capability to accurately recommend excipients and thereby accelerate the formulation development pipeline. While conventional metrics such as accuracy, precision, and F1 score can be applied, these deterministic measures are not well aligned with the inherently probabilistic and generative nature of language model outputs. Consequently, future work should focus on defining and validating domain-aware evaluation metrics, like VELVET, that enable rapid and reliable screening of language models for suitability in pharmaceutical formulation tasks.

We also observed instances of *catastrophic forgetting*, a phenomenon in which previously learned representations are overwritten during subsequent training phases [43]. We hypothesise that this mechanism underlies the generation of nonsensical alphanumeric outputs by certain LLMs. Our analysis indicates that smaller-capacity models are more susceptible to catastrophic forgetting, a trend that is consistent with established neural scaling laws [44]. Prior work has shown that fine-tuning LLMs on data from a different languages can induce catastrophic forgetting [45]. We suspect a similar effect is occurring in this setting, as domain-specific terms such as *Eudragit*, *Klucel ELF* and *PEG8000* are unlikely to appear frequently in the base training corpora of these models. This behaviour was unexpected in the case of BioGPT, which has previously demonstrated strong performance on biomedical tasks involving API nomenclature [34, 46]. Catastrophic forgetting is a well-characterised issue in the LLM literature, and multiple mitigation strategies have been proposed [47]. Should these approaches prove insufficient, an alternative strategy would be to explicitly train or adapt language models using corpora enriched with excipient-specific terminology.

Furthermore, this study demonstrates that selecting an LLM for PEFT based solely on domain-specific pretraining does not necessarily lead to optimal performance. Although BioGPT was pretrained on approximately 115 million PubMed articles and has been shown to outperform other biomedical language models on a range of tasks [46], these results indicate that model architecture and training dynamics must also be considered. In particular, architectural characteristics may play a decisive role in downstream performance for pharmaceutical formulation tasks. While the development of a formulation-specific LLM may ultimately be warranted, immediate research efforts should prioritise a systematic investigation of fine-tuning strategies and their effects before undertaking the substantially more computationally intensive process of training a domain-specific model from scratch.

Regarding learning rate, previous work in deep learning models for pharmaceutical applications has highlighted the significance of choosing the correct learning rate [13, 40]. Herein, we confirmed that a learning rate of either $10^{-2}$ or $10^{-6}$ can lead to all four LLMs producing nonsensical outputs. However, one limitation is that we did not explore the combination of longer epochs with these learning rate. Hence, future work should also investigate a wider range of fine-tuning parameter combinations, computational resources permitting.

Ultimately, an AI system capable of accurately recommending pharmaceutical excipients while simultaneously supporting a broader range of tasks is preferable to traditional narrowly scoped models with limited generalisation capacity. As pharmaceutical science continues to evolve, the ability to adapt to new tasks and knowledge domains becomes increasingly important. Currently, there are a number of proposed AI tools covering different functions, such as predicting processing conditions, product mechanical properties and PKPD properties [48-50], however, a unified AI platform that covers all aspects of the pharmaceutical pipeline would be more pragmatic.

# 5 Conclusion

The study investigated the potential of fine-tuning LLMs for generating excipient recommendations given an API and its proportion, thereby accelerating 3D printing formulation development. Four different LLMs were investigated, including one trained on biomedical data that was hypothesised to yield better results. The outcome contradicted our hypothesis, and found that model architecture was more important. We investigated both fine tuning and generative parameter optimisation and revealed both can impact model performance. Furthermore, we highlighted that standard metrics do not necessarily correlate to useful excipient recommendations, which led us to propose the need for a new metric, such that model performance can be gauged by their ability to generate sensical whilst pharmaceutically-relevant responses. Nonetheless, the results are promising and demonstrate that with additional investigations, the pharmaceutical community has the potential to leverage these powerful AI technologies, ultimately addressing current developmental bottlenecks.

# 6 Appendices

| Appendix Table A1 shows the software packages used for each type of analysis; the packages in bold have been used throughout the study. | | |
|---|---|---|
| Analysis | Software packages used | Versions |
| Exploratory data analysis | pandas<br>numpy<br>matplotlib<br>seaborn | 2.2.2<br>2.0.2<br>3.10.0<br>0.13.2 |
| Training and Generation | transformers<br>datasets<br>peft<br>torch<br>bitsandbytes<br>accelerate<br>rouge-score<br>nltk,nltk(“punkt”),nlkt(“punkt_tab”)<br>unsloth<br>json<br>time<br>torch<br>os<br>shutil<br>sys<br>google.colab | 4.57.3<br>4.0.0<br>0.18.0<br>2.7.0+cu126<br>0.49.0<br>1.12.0<br><br>0.1.2<br>3.9.1<br>2025.12.8<br>Python 3.12.12<br>Python 3.12.12<br>Python 3.12.12<br>Python 3.12.12<br>Python 3.12.12<br>Python 3.12.12<br>1.0.0 |
| Evaluation | rouge_score<br>sacrebleu<br>numpy<br>matplotlib<br>pandas<br>sacremoses<br>google.colab<br>os | 0.1.2<br>2.5.1<br>2.0.2<br>3.10.0<br>2.2.2<br>0.1.1<br>1.0.0<br>Python 3.12.12 |

| | **sys** | Python 3.12.12 |
|---|---|---|
| Statistical evaluation | sacremoses<br>sacrebleu<br>rouge_score<br>re<br>statsmodels<br>scipy<br>google.colab<br>os<br>sys | 0.1.1<br>2.5.1<br>0.1.2<br>Python 3.12.12<br>0.14.6<br>1.16.3<br>1.0.0<br>Python 3.12.12<br>Python 3.12.12 |